%% file: main.tex
\documentclass[preprint,12pt,authoryear]{elsarticle}

\newif\ifanon
\anonfalse

\usepackage{amssymb}
\usepackage{amsmath}
\usepackage{amsthm}
\usepackage{algorithm}
\usepackage{algpseudocode}
\usepackage{multirow}
\usepackage{graphicx}
\usepackage{booktabs}
\usepackage{url}
\usepackage[table]{xcolor}
\usepackage[hidelinks]{hyperref}
\ifanon\hypersetup{pdfauthor={},pdfsubject={},pdfkeywords={}}\fi
\graphicspath{{figures/}}
\newtheorem{definition}{Definition}

\input{tables/numbers}

\newcommand{\eprec}{\varepsilon_{\mathrm{prec}}}
\newcommand{\ecov}{\varepsilon_{\mathrm{cov}}}

\journal{Expert Systems with Applications}

\begin{document}

\begin{frontmatter}

\title{RCProb: Probabilistic rule extraction from classification tree ensembles}

\ifanon
\author{}
\else
\author{Josue Obregon}
\ead{jobregon@seoultech.ac.kr}

\affiliation{organization={Department of Industrial Engineering, Seoul
National University of Science and Technology (SeoulTech)},
            city={Seoul},
            postcode={01811},
            country={Republic of Korea}}
\fi

\begin{abstract}
Tree ensembles provide strong classification performance but usually behave
as black-box models. Post-hoc interpretability techniques such as RuleCOSI+
extract a small ruleset that approximates the ensemble, but this
simplification can leave the probabilities attached to the extracted rules
unreliable. In particular, RuleCOSI+ assigns empirical class probabilities to
the extracted rules and repeatedly uses those rule statistics during its
greedy combination and simplification procedure. We present RCProb, a
probabilistic extension that uses smoothed atomic class-conditional evidence
for the expensive search stages and a support-adaptive mixture with an ensemble-informed m-estimate for the final rule probabilities. The method is
evaluated on 18 binary and 5 multiclass datasets using random forest (RF) and
gradient boosting machine (GBM) ensembles. Relative to RuleCOSI+, the median
paired log-loss reduction is 71.9\% for RF and 62.5\% for GBM, with both
differences remaining significant after Holm correction. The number of rules
decreases by 38.7\% for RF and 38.5\% for GBM, while the primary tests do not
detect a macro-F1 difference. Confidence-ECE also decreases for both
ensembles, with statistical support for RF after correction. A separate
controlled experiment with dedicated calibration data shows that native
RCProb probabilities are competitive with RuleCOSI+ followed by temperature
scaling, although additional post-hoc calibration can still improve RCProb.
The results show that probability estimation is an important part of rule
extraction and not only a post-processing step.
\end{abstract}


\end{frontmatter}

\input{sections/01_intro}
\input{sections/02_background}
\input{sections/03_preliminaries}
\input{sections/04_method}
\input{sections/05_experiments}
\input{sections/06_results}
\input{sections/07_discussion}

\input{sections/08_conclusion}

\ifanon\else
\section*{CRediT authorship contribution statement}
\textbf{Josue Obregon:} Conceptualization, Methodology, Software, Formal
analysis, Investigation, Data curation, Writing -- original draft, Writing
-- review \& editing, Visualization, Funding acquisition.

\section*{Declaration of competing interest}
The author declares that he has no known competing financial interests or
personal relationships that could have appeared to influence the work
reported in this paper.
\fi

\section*{Data availability}
All datasets used in this study are publicly available benchmark datasets.

\ifanon\else
\section*{Acknowledgments}
This study was supported by the Research Program funded by SeoulTech (Seoul
National University of Science and Technology).
\fi

\appendix
%% The supplementary material travels with this preprint as an appendix.
%% Tables keep the S numbering the main text points to, so every literal
%% "Table~S<n>" pointer resolves inside this one document.
\setcounter{table}{0}
\renewcommand{\thetable}{S\arabic{table}}

\section{Notation differences from RuleCOSI+}
The main manuscript follows the notation of RuleCOSI+ where practical, citing
that paper in its reference list. Three symbols were written differently for
readability, and Table~S1 gives the correspondence. All other symbols keep the
meaning they have in that paper.

\begin{table}[htbp]
\centering
\caption{Symbols written differently in this manuscript.}
\label{tab:supp-notation}
\small
\resizebox{\textwidth}{!}{%
\begin{tabular}{lll}
\toprule
This manuscript & RuleCOSI+ & Quantity \\
\midrule
$N_r$ & $n_r$ & Number of training records satisfying the antecedent \\
$\mathrm{prec}(r)$ & $\mathrm{conf}(r)$ & Precision of a rule on the training data \\
$\varepsilon_{\mathrm{prec}}$, $\varepsilon_{\mathrm{cov}}$ & $\alpha$, $\beta$ & Rule precision and coverage thresholds \\
\bottomrule
\end{tabular}}
\end{table}

RuleCOSI+ calls this measure the confidence of a rule. We use precision
because the main manuscript has to distinguish it from the probability
$\widehat{p}(y\mid r)$ that a rule states. RuleCOSI+ needs one symbol, since
there the stated probability is the empirical precision.

\section{Dataset-level primary results}
The main manuscript reports RF and GBM separately. Tables~S2 and~S3 give the
underlying dataset-level means for the primary RC+--RCProb comparison.
\input{tables/s1_per_dataset_rf}
\input{tables/s1_per_dataset_gbm}

\section{Secondary five-method comparisons}
The primary RCProb--RC+ tests are reported in the main manuscript.
Tables~S4 and~S5 give the secondary five-method Friedman and post-hoc
comparisons for RF and GBM, and Tables~S6 and~S7 the corresponding average
ranks.
\input{tables/s10_secondary_rf}
\input{tables/s10_secondary_gbm}
\input{tables/s11_ranks_rf}
\input{tables/s11_ranks_gbm}

\section{Controlled post-hoc calibration experiment}
Table~S8 reports the complete descriptive comparison by base ensemble,
including the higher-capacity Dirichlet-L2 sensitivity analysis and post-hoc
calibration applied to RCProb. The six-dataset experiment does not support a
claim of statistical superiority.
\input{tables/s2_calibration_full}

\section{Rule-level reliability diagnostics}
Table~S9 reports the rule-level quantities. These are descriptive
applications of calibration-in-the-large within rule-defined held-out
subgroups, and are not proposed as a new scalar calibration metric.
\input{tables/s3_rule_diagnostics}

\section{Aggregate-inference sensitivity}
The main manuscript uses decision-list inference. Table~S10 repeats the
RC+--RCProb comparison under aggregate inference separately for RF and GBM.
\input{tables/s5_aggregate}

\section{Held-out use of overlapping rules}
Table~S11 separates antecedent matching from first-rule selection under the
standardized decision-list representation. Table~S12 then compares
decision-list and aggregate predictions only where both modes selected
exactly the same stored ruleset.
\input{tables/s6_rule_use}
\input{tables/s7_identical_inference}

\section{Targeted parameter sensitivity}
The main benchmark fixes the smoothing strength $\eta=1$, the shrinkage
strength $\tau=5$, and the mixing support $n_0=50$, using the notation of the
main manuscript. The additive smoothing value $\eta=1$ is the standard Laplace
choice. Table~S13 reports the targeted support-mixing and shrinkage analyses
separately for RF and GBM.

In Panel A, increasing $n_0$ from 50 to 400 progressively increases log-loss
while F1 and rule count remain unchanged, so the study gives no evidence that
larger support-transition values improve the tested tradeoff. In Panel B, the
region $\tau\in[0,5]$ is nearly flat, with small ensemble-dependent
differences, and clearer degradation appears at $\tau\geq20$. We therefore
treat $\tau=5$ as a fixed moderate setting and make no claim that it is
empirically optimal.

Panel B was run under aggregate inference and is interpreted as a separate
reduced sensitivity check, not as a direct reproduction of the primary
decision-list benchmark. Both panels use fixed ensemble settings and assess
stability of the reported behavior under changes to these two parameters.
\input{tables/s8_sensitivity}

\bibliographystyle{elsarticle-harv}
\bibliography{2026_RCProb}

\end{document}

%% file: tables/numbers.tex
\newcommand{\NDatasets}{23}
\newcommand{\NFolds}{15}

%% file: sections/01_intro.tex
\section{Introduction}
\label{sec:intro}

Tree ensembles such as random forests (RF) and gradient boosting machines
(GBM) remain strong predictors for tabular classification problems, but
usually, their large collections of trees are difficult to inspect directly.
Post-hoc rule extraction addresses this problem by replacing the ensemble
with a smaller set of IF--THEN rules that approximates its decisions
\citep{barredoarrietaXAI2020}. Rule-based explanations are attractive
because the same representation can support a global view of the model and a
local explanation for an individual prediction.

A rule can communicate more than a class label. When a rule carries a class
probability vector, it can state that a class is predicted with a
probability of 0.95 for example, instead of simply returning the class. Such
a number is useful only if it has a meaningful probabilistic interpretation.
Calibration requires predicted probabilities to agree with observed
frequencies over comparable cases, while proper scoring rules also measure
the overall quality of the complete probability vector
\citep{silva2023classifier,gneitingStrictlyProperScoring2007}. These
properties are separate from classification accuracy or fidelity, so a
ruleset may reproduce the hard decisions of its source ensemble while
providing poor probability estimates.

The distinction matters wherever the decision, and not only the label, is
the output. Calibrated probabilities allow a decision threshold to be
adjusted when class priors or misclassification costs change, without
relearning the explanatory ruleset \citep{silva2023classifier}. In consumer
credit scoring, default probabilities carry information that a good or bad
classification discards, and they let a lender set a cutoff that reflects
its own cost of default \citep{kruppaConsumerCreditRisk2013}. In gene
expression classification, Naive Bayes probability estimates degrade when
the sample is small relative to the number of measured features, which is
the same small-count problem that arises for a sparsely supported rule
\citep{chandraRobustApproachEstimating2011}. Recent work on calibrated
explanations argues that an explanation should carry its uncertainty, so
that a user can judge how far to rely on it
\citep{lofstromCalibratedExplanations2024}.

Post-hoc simplification of tree ensembles is an established line of work.
RuleFit \citep{Friedman2008} converts tree leaves into features and fits a
sparse linear model over them. inTrees \citep{Deng2019} measures the rules
of the ensemble by frequency, error, and length, prunes conditions whose
removal leaves the error nearly unchanged, and selects a compact and
non-redundant subset. DefragTrees (DFT) \citep{haraMakingTreeEnsembles2018}
formulates simplification as Bayesian model selection over a partition of
the feature space, and TERM \citep{zhuTERMTreeEnsemble2025} covers the
ensemble with a compact subset of its rules. RuleCOSI \citep{Obregon2019}
introduced a greedy combination and simplification procedure for boosted
ensembles under class imbalance, and RuleCOSI+ (RC+)
\citep{obregonRuleCOSIRuleExtraction2023} generalized it to independent and
dependent classification tree ensembles. RC+ extracts the rules of every
tree, combines rulesets pairwise, prunes them by sequential covering, and
generalizes the survivors by removing conditions under a pessimistic error
estimate. These methods pursue the same objective and differ in what the
simplified model reports. RuleFit predicts through a linear combination, so
no single rule carries a probability of its own, whereas DFT is
probabilistic by construction and assigns a class distribution to each
region. RC+ reports a class probability vector with the rule that fires,
which makes that number part of the explanation the user sees. Two further
lines of work bear on the same number without simplifying an ensemble.
Probabilistic rule learners \citep{wangBayesianFrameworkLearning2017} model
uncertainty directly but build a classifier from data, and well-established
post-hoc calibration
\citep{platt1999probabilistic,guoCalibrationModernNeural2017} repairs the
probabilities of an already fitted model once the ruleset has been selected.

In particular, the probabilities that RC+ reports are produced by the same
greedy search that selects the rules, and this is where the gap lies.
Empirical coverage and precision are evaluated repeatedly while rules are
combined, pruned, and generalized. Sparse candidates can obtain extreme
empirical proportions from only a few observations, and data-driven
selection can favor estimates that are unusually high because of sampling
variation. This is the general post-selection phenomenon often described as
a winner's curse \citep{andrewsInferenceWinners2024}, and it appears in
induction algorithms as a multiple comparisons effect that biases the score
of the selected candidate upward
\citep{jensenMultipleComparisonsInduction2000}. Correcting overly optimistic
rule probabilities is the same problem as calibration, and
\citet{furnkranzFoundationsRuleLearning2012} note that it has received
considerably less attention in rule learning than in decision tree learning.
The measures normally used to compare simplified rulesets, such as
predictive performance, fidelity, and model size, do not assess the quality
of a probability estimate \citep{silva2023classifier}, so the values
attached to the final rules can degrade without appearing in the comparison.
Recent work connecting calibration and explanation also indicates that
uncertainty information can materially affect how an explanation is
interpreted \citep{lofstromCalibratedExplanations2024}.

We present RCProb, a probabilistic extension of RC+ for classification tree
ensembles. RCProb retains the extract, combine, and simplify structure of
its predecessor. It replaces repeated empirical probability calculations in
the expensive search stages with smoothed atomic class-conditional evidence
combined through a Naive Bayes approximation. For rules whose exact support
is recomputed, a support-adaptive mixture combines that estimate with an
ensemble-informed m-estimate. Sparse rules consequently rely more on the
local probability information inherited from the ensemble, while the atomic
evidence receives more weight as support grows.

The evaluation uses 18 binary and 5 multiclass datasets with RF and GBM
ensembles. RCProb is compared directly with RC+ under identical outer splits
and base ensembles, and is also placed in context with inTrees, DFT, and a
calibrated single decision tree. The probability outputs are evaluated with
log-loss and the Brier score as proper scoring rules, together with
confidence-ECE and classwise-ECE. A separate controlled experiment compares
native RCProb probabilities with RC+ followed by post-hoc temperature
scaling using a dedicated calibration partition. Relative to RC+, RCProb
reduces the median paired log-loss by 71.9\% for RF and 62.5\% for GBM and
produces 38.7\% and 38.5\% fewer rules respectively, while the primary tests
do not detect a macro-F1 difference.

The main contributions are as follows.

\begin{itemize}
\item We develop RCProb, a probabilistic rule extraction method that uses
smoothed atomic evidence during rule search and a support-adaptive hybrid
probability for the final rules. The method applies to both independent and
dependent classification tree ensembles supported by RC+.

\item We show that the probability estimates produced during extraction are
a substantive part of the simplification problem. Replacing them lowers
log-loss and reduces the number of rules in both base ensembles after Holm
correction, while macro-F1 is not distinguished, so the probability gain does
not come at a measured cost in hard classification.

\item RCProb produces competitive rule probabilities without a calibration
step, so no held-out calibration partition is required. In a controlled 
experiment with a subset of six representative datasets, RCProb is 
better than temperature-scaled RC+ on every probability measure, 
and calibration still improves RCProb further.

\end{itemize}

The remainder of the paper is organized as follows.
Section~\ref{sec:related} reviews related work. Section~\ref{sec:prelim}
introduces the notation and problem setting. Section~\ref{sec:method}
describes RCProb, and Section~\ref{sec:experiments} presents the
experimental design. Sections~\ref{sec:results} and~\ref{sec:discussion}
report and discuss the results, followed by the conclusion in
Section~\ref{sec:conclusion}.

%% file: sections/02_background.tex
\section{Related work}
\label{sec:related}

This section reviews the work on which RCProb builds. We first discuss
post-hoc simplification of tree ensembles, followed by probability
estimation in trees and rule models, and classifier calibration.

\subsection{Rule extraction and simplification of tree ensembles}
\label{sec:rw-extraction}

Post-hoc simplification methods transform a trained ensemble into a
transparent model that approximates it. RuleFit \citep{Friedman2008}
converts tree leaves into binary features and fits a sparse linear model
over them. The inTrees framework \citep{Deng2019} extracts rules from an
ensemble and measures them by frequency, error, and length. Conditions are
pruned individually when the relative increase in error caused by removing a
condition falls below a threshold, and a compact non-redundant subset is
then obtained by regularized feature selection over the surviving
conditions, which penalizes longer conditions. The outcome of each rule is
the most frequent class among the instances that satisfy its condition. DFT
\citep{haraMakingTreeEnsembles2018} formulates simplification as Bayesian
model selection over a partition of the feature space. Forest-based trees
\citep{sagiExplainableDecisionForest2020,
sagiApproximatingXGBoostInterpretable2021} and born-again tree ensembles
\citep{vidalBornAgainTreeEnsembles2020} return a single surrogate tree
instead of a ruleset. TERM \citep{zhuTERMTreeEnsemble2025} uses an adaptive
set cover algorithm to select a compact subset of ensemble rules for binary
classification.

Other studies formulate rule simplification as an explicit optimization
problem. The two-stage method of \citet{dongTwostageRuleExtraction2021}
first simplifies rules locally and then performs global selection with a
multi-objective genetic algorithm. \citet{maissaeForestOREMiningOptimal2024}
pose selection as a mixed-integer program balancing coverage, overlap and
model size. \citet{gulowatyExtractingInterpretableDecision2021} represent
pairwise rule overlap as a graph and select mutually non-overlapping rules,
whereas \citet{mashayekhiRuleExtractionDecision2017} combine hill climbing
with sparse group lasso. \citet{takemuraGeneratingExplainableRule2021} cast
rule extraction as answer set programming with global and pairwise
constraints.

RuleCOSI \citep{Obregon2019} introduced a greedy combination and
simplification framework for boosted ensembles under class imbalance. RC+
\citep{obregonRuleCOSIRuleExtraction2023} generalized the framework to
independent and dependent classification tree ensembles and made the
combination procedure practical for larger ensembles. Its main stages are
extraction of raw tree rules, iterative ruleset combination, sequential
covering pruning, and condition generalization under a pessimistic error
estimate.

These methods mainly address which rules should represent the ensemble and
how compact or faithful the resulting transparent model should be. The
present study addresses the reliability of the class probabilities attached
to extracted rules. This issue is particularly relevant to RC+, whose
decision-list rules carry a class probability vector derived from the
training records covered by the rule.

\subsection{Probability estimation in trees and rule models}
\label{sec:rw-probabilistic}

In rule learning, the weight attached to a rule is commonly the relative
frequency of its class among the covered records, which overestimates the
true probability because of overfitting, so Laplace-corrected and
m-estimates are the customary correction
\citep{furnkranzFoundationsRuleLearning2012}. Probability smoothing in tree
models is established practice for the same reason, because raw leaf
proportions can be unstable when terminal nodes contain few observations.
\citet{provostTreeInductionProbability2003} show that a leaf holding five
instances of one class yields a maximum likelihood estimate of 1.0, and
apply the Laplace correction to soften such estimates toward a uniform
prior. Recent work has continued to study post-hoc smoothing of tree
probabilities, including Beta-Binomial smoothing of random-forest leaves
\citep{pfeiferTreeSmoothingPosthoc2025}. Such procedures modify probability
estimates after the predictive structure has been learned.

Probabilistic rule learners take a different approach. For example,
\citet{wangBayesianFrameworkLearning2017} place priors over rule structure
and infer compact disjunctive models through Bayesian model selection. These
methods produce probabilistic rule classifiers directly from data, whereas
RCProb is designed to simplify an already trained tree ensemble.

The estimator used by RCProb also relates to smoothed probability estimation
in Naive Bayes models. \citet{chandraRobustApproachEstimating2011} study the
instability of maximum-likelihood class-conditional estimates when counts
are small and use prior information to stabilize them. RCProb applies the
same principle inside the extraction search, so that probability decisions
are not made from small empirical counts alone.

\subsection{Classifier calibration}
\label{sec:rw-calibration}

A probabilistic classifier is calibrated when its stated probabilities agree
with the frequencies of the corresponding events
\citep{silva2023classifier}. Binned calibration measures depend on how the
predictions are partitioned, and a low calibration error alone does not
guarantee useful probabilistic predictions
\citep{kullTemperatureScalingObtaining2019,nixonMeasuringCalibrationDeep2020}.
Proper scoring rules assess the complete probabilistic prediction and are
therefore complementary \citep{gneitingStrictlyProperScoring2007}.
Section~\ref{sec:measures} defines the measures used in this study.

Post-hoc calibration learns a transformation from held-out predictions.
Platt scaling \citep{platt1999probabilistic}, isotonic regression
\citep{zadrozny2002transforming}, Beta calibration
\citep{kullBetaCalibrationWellfounded2017}, temperature scaling
\citep{guoCalibrationModernNeural2017}, and Dirichlet calibration
\citep{kullTemperatureScalingObtaining2019} are established examples. A
post-hoc map can substantially improve the probability vector output by an
already fitted model. It operates after the ruleset has been selected, so it
leaves the extraction search and the probability estimates that guided it
unchanged, and it requires held-out data on which the calibration map can be
fitted. Section~\ref{sec:ex-posthoc} evaluates post-hoc calibration in a
separate experiment with a dedicated calibration partition.

Calibration can also be considered within identifiable subgroups. The
multicalibration framework requires predicted probabilities to agree with
observed frequencies across specified subpopulations
\citep{hebertjohnsonMulticalibration2018}. A decision-list rule naturally
defines such a subgroup because all held-out instances assigned to that rule
receive the same rule probability vector. We use this principle only as a
descriptive rule-level diagnostic; formal probabilistic comparisons use the
standard measures defined in Section~\ref{sec:measures}.

%% file: sections/03_preliminaries.tex
\section{Preliminaries}
\label{sec:prelim}

This section introduces the notation used throughout the paper. We define
the classification setting, decision rules and rulesets, and the rule
statistics used by RCProb. Table~\ref{tab:notation} collects the symbols for
reference. The notation follows RC+
\citep{obregonRuleCOSIRuleExtraction2023} where practical. Some symbols were
adapted for readability, and Table~S1 lists the differences.

\input{tables/t0_notation}

\subsection{Classification tree ensembles}
\label{sec:pre-ensembles}

Let $D = \{(x_i, y_i)\}$ with $i = 1$ to $n$ be a training set, where $x_i
\in \mathbb{R}^m$ is a vector of $m$ attributes and $y_i \in \mathcal{Y} =
\{y_1, \dots, y_C\}$ is a class label. We consider classification problems
with two or more classes.

A classification tree ensemble $H$ is a set of trees $h_k \in H$ for $k = 1$
to $K$, each constructed from $D$. The trees may be built independently, as
in bagging and random forests, or dependently, as in gradient boosting
machines (GBM). When $H$ classifies an instance it aggregates the outputs of
its trees through a function $\phi$, so that $H(x) = \phi(h_1(x), \dots,
h_K(x))$. For independent ensembles the output of a tree is a class
probability vector and $\phi$ is their average. For dependent ensembles the
output is a score, $\phi$ is their sum, and the result is transformed to a
probability vector by a logistic or softmax function.

\subsection{Decision rules and rulesets}
\label{sec:pre-rules}

\begin{definition}[Decision rule]
\label{def:rule}
A decision rule $r$ has the form $r : A \rightarrow y$, where $A$ is the
antecedent, a conjunction of atomic conditions $a \in A$, and $y$ is the
consequent class. An atomic condition has the form $x_j \le t$ or $x_j >
t$ for an attribute index $j$ and a threshold $t$.
\end{definition}

Following \citet{obregonRuleCOSIRuleExtraction2023}, we write the antecedent
of a rule $r$ as $r.A$, its consequent as $r.y$, and the class probability
vector attached to it as $r.p$. The antecedent encloses a region $\pi_r$ of
the feature space, and the rule is said to match (or fire on) a record $x$
when $x \in \pi_r$.

Rules are grouped into a ruleset $R$, and $R^{\otimes}$ denotes the
simplified ruleset that an extraction method returns. We consider two
inference modes. In
a decision list the rules are evaluated in order and the first matching
rule is selected. Its consequent and probability vector determine the
prediction. Under aggregate inference every matching non-default rule
contributes, and the prediction is obtained by
normalizing the sum of their probability vectors,
\begin{equation}
\label{eq:aggregate}
p(x) = \frac{\sum_{r \in R(x)} r.p}
{\left\lVert \sum_{r \in R(x)} r.p \right\rVert_1},
\qquad R(x) = \{ r \in R : x \in \pi_r \}.
\end{equation}
A ruleset is completed by a default rule $r_d : (\,) \rightarrow y_d$
whose antecedent is empty. It is selected when no non-default rule matches the record,
and $y_d$ is the majority class among the training records the remaining
rules leave uncovered.

\subsection{Rule support, coverage, and precision}
\label{sec:pre-measures}

For a rule $r$, let
\begin{equation}
D_r = \{(x_i,y_i) \in D : x_i \in \pi_r\}
\end{equation}
be the training records satisfying its antecedent. We denote the support
count by $N_r = |D_r|$ and the number of supported records from class $y$
by $N_{r,y}$. The coverage definition used in RC+ is retained,
\begin{equation}
\label{eq:coverage}
\mathrm{cov}(r) = \frac{N_r}{n},
\end{equation}
and its precision on the training data is
\begin{equation}
\label{eq:confidence}
\mathrm{prec}(r) = \frac{N_{r,r.y}}{N_r}.
\end{equation}
The same rule metric appears in the literature as precision, as rule accuracy,
and as confidence in association rule mining
\citep{furnkranzFoundationsRuleLearning2012}. RC+ uses the last name; we use
the first.

The distinction between $N_r$ and $\mathrm{cov}(r)$ is important in
Section~\ref{sec:m-mixture}, because the support-adaptive mixture is
parameterized in numbers of records, not fractions of the training set.

The probability vector attached to a rule need not be identical to its
empirical training distribution. We write $r.p_{r.y}$ for the entry of $r.p$
that corresponds to the class the rule predicts. In RC+, $r.p$ is obtained
from empirical training frequencies, so $r.p_{r.y}=\mathrm{prec}(r)$. In
RCProb, $r.p$ is the posterior estimate defined in Section~\ref{sec:method},
and the two differ.

A rule also defines a natural subgroup on held-out data. Under decision-list
inference, let $T_r$ be the held-out records assigned to $r$ after all
preceding rules have been applied, let $M_r = |T_r|$, and let $M_{r,y}$ be
the number of those records whose class is $y$. The precision of the rule on
this subgroup, $\mathrm{prec}_{\mathrm{test}}(r) = M_{r,r.y}/M_r$, is
Equation~\ref{eq:confidence} evaluated on held-out records instead of
training records.

Calibration-in-the-large compares an average predicted probability with
the corresponding observed event rate \citep{vanCalsterCalibrationAchilles2019}.
Because every record assigned to one decision-list rule receives the same
probability $r.p_{r.y}$ for the event $Y=r.y$, applying this standard
calibration concept within the rule-defined subgroup reduces directly to the
gap
\begin{equation}
\label{eq:rulegap}
g_r = r.p_{r.y} - \mathrm{prec}_{\mathrm{test}}(r).
\end{equation}
Positive values indicate overconfidence and negative values indicate
underconfidence. This is also consistent with the broader subgroup
calibration principle studied by \citet{hebertjohnsonMulticalibration2018}.
The hierarchy of \citet{vanCalsterCalibrationAchilles2019} places
calibration-in-the-large at its first level and, at its highest, strong
calibration, which requires agreement for every combination of predictor
values and which that work describes as a utopic goal. Grouping by rule falls
between these levels, because a rule fixes a region of the feature space. We
therefore use $g_r$ as a descriptive within-rule difference and not as
evidence of calibration at a particular level of that hierarchy. Rules with $M_r=0$ have no
observable held-out precision in that fold and are excluded from this
diagnostic.

The practical problem is most visible for sparse rules. A rule supported by
only a few training records can obtain $\mathrm{prec}(r)=1$ even though the
corresponding region has substantial uncertainty. A greedy search also
selects rules using this observed quantity, so unusually favorable training
proportions can influence which candidates survive. RCProb replaces the
probability estimate used during the search stages and the probability
vector attached to the final rules.

\subsection{Problem statement}
\label{sec:pre-problem}

\begin{definition}[Probabilistic post-hoc rule extraction]
\label{def:extraction}
Given a trained ensemble $H$ and a training set $D$, a probabilistic rule
extraction function $G$ produces a ruleset $R = G(H, D)$ such that, for every
$x \in \mathbb{R}^m$,
\begin{align}
\label{eq:fidelity}
R(x) &\approx H(x), \\
\label{eq:probtarget}
r.p &\approx P(y \mid x \in \pi_r),
\end{align}
where the first line is agreement on the predicted class, $r$ is the rule
that produces the prediction for $x$, and $P(y \mid x \in \pi_r)$ is the
conditional class distribution of the region enclosed by $r$.
\end{definition}

Equation~\ref{eq:fidelity} is the standard post-hoc extraction requirement.
Equation~\ref{eq:probtarget} is added here, and it is assessed on held-out
records with proper scoring rules, which are minimized by the true
conditional distribution \citep{gneitingStrictlyProperScoring2007}, together
with calibration measures, which compare stated probabilities with observed
frequencies \citep{silva2023classifier}.

%% file: tables/t0_notation.tex
\begin{table*}[htbp]
\centering
\caption{Notation used throughout the paper.}
\label{tab:notation}
\footnotesize
\setlength{\tabcolsep}{4pt}
\renewcommand{\arraystretch}{0.95}
\begin{minipage}[t]{0.485\textwidth}
\vspace{0pt}
\begin{tabular}{@{}l p{0.63\linewidth}@{}}
\toprule
Symbol & Meaning \\
\midrule
\multicolumn{2}{@{}l}{\textit{Data and ensemble}} \\
$D$ & Training set $\{(x_i,y_i)\}$, $i=1$ to $n$ \\
$n$ & Number of training records \\
$m$ & Number of attributes \\
$C$ & Number of classes \\
$\mathcal{Y}$ & Set of class labels \\
$H$, $h_k$, $K$ & Ensemble, its $k$th tree, number of trees \\
$d$ & Maximum tree depth \\
$N_{\mathrm{test}}$ & Number of held-out records in a fold \\
\addlinespace
\multicolumn{2}{@{}l}{\textit{Rules and rulesets}} \\
$r: A \rightarrow y$ & Decision rule (Definition~\ref{def:rule}) \\
$r.A$, $r.y$, $r.p$ & Its antecedent, consequent class, and class
probability vector \\
$r.p_{r.y}$ & Entry of $r.p$ for the predicted class \\
$a$ & Atomic condition in an antecedent \\
$\pi_r$ & Feature-space region enclosed by $r.A$ \\
$A$, $A^*$ & Set and multiset of the conditions in a ruleset \\
$P(y\mid x\in\pi_r)$ & True class distribution of that region \\
$R$, $R^{\otimes}$ & Ruleset, and the simplified ruleset returned \\
$r_d$ & Default rule, used when no other rule matches \\
\addlinespace
\multicolumn{2}{@{}l}{\textit{Simplification thresholds}} \\
$\eprec$ & Rule precision threshold \\
$\ecov$ & Rule coverage threshold \\
$c$ & Confidence level of the pessimistic error estimate
(Equation~\ref{eq:pessimistic}) \\
\bottomrule
\end{tabular}
\end{minipage}
\hfill
\begin{minipage}[t]{0.485\textwidth}
\vspace{0pt}
\begin{tabular}{@{}l p{0.63\linewidth}@{}}
\toprule
Symbol & Meaning \\
\midrule
\multicolumn{2}{@{}l}{\textit{Rule statistics on the training data}} \\
$N_r$ & Records satisfying $r.A$ \\
$N_{r,y}$ & Records satisfying $r.A$ whose class is $y$ \\
$N_y$ & Training records of class $y$ \\
$\mathrm{cov}(r)$ & Coverage of a rule $r$ (Equation~\ref{eq:coverage}) \\
$\mathrm{prec}(r)$ & Precision of a rule on the training data
(Equation~\ref{eq:confidence}) \\

\addlinespace
\multicolumn{2}{@{}l}{\textit{Held-out rule diagnostics}} \\
$M_r$ & Held-out records assigned to $r$ \\
$M_{r,y}$ & Those of class $y$ \\
$\mathrm{prec}_{\mathrm{test}}(r)$ & Precision on those records \\
$g_r$ & Stated probability minus held-out precision, positive when
overconfident (Equation~\ref{eq:rulegap}) \\
\addlinespace
\multicolumn{2}{@{}l}{\textit{RCProb estimators and parameters}} \\
$\eta$ & Additive smoothing strength \\
$p(y)$, $p(a\mid y)$ & Smoothed class prior and atomic likelihood
(Equations~\ref{eq:prior} and~\ref{eq:likelihood}) \\
$p_{\mathrm{NB}}(y\mid r)$ & Naive Bayes estimate used during the search
(Equation~\ref{eq:nb}) \\
$\mathbf{q}_r$ & Local class distribution from the ensemble \\
$\tau$, $\tau_r$ & Prior strength and its per-rule cap \\
$\widetilde{p}(y\mid r)$ & Ensemble-informed m-estimate
(Equation~\ref{eq:leaf}) \\
$n_0$ & Support at which the two estimates weigh equally \\
$\lambda(r)$ & Weight on the Naive Bayes estimate
(Equation~\ref{eq:lambda}) \\
$\widehat{p}(y\mid r)$ & Hybrid probability attached to the final rule
(Equation~\ref{eq:hybrid}) \\
\bottomrule
\end{tabular}
\end{minipage}
\end{table*}

%% file: sections/04_method.tex
\section{RCProb: probability estimation for extracted rules}
\label{sec:method}

RCProb retains the extract, combine, and simplify structure of RC+ but
changes how probability information is estimated during the search and
attached to the final rules. The method has three probabilistic components:
a Naive Bayes estimate over atomic conditions, an ensemble-informed
m-estimate based on exact rule counts, and a support-adaptive mixture of the
two. The Naive Bayes estimate is used in the merge and generalization
searches, whereas the hybrid estimate is computed for rules whose exact
support has been recomputed.

\begin{figure*}[htbp]
\centering
\includegraphics[width=\textwidth]{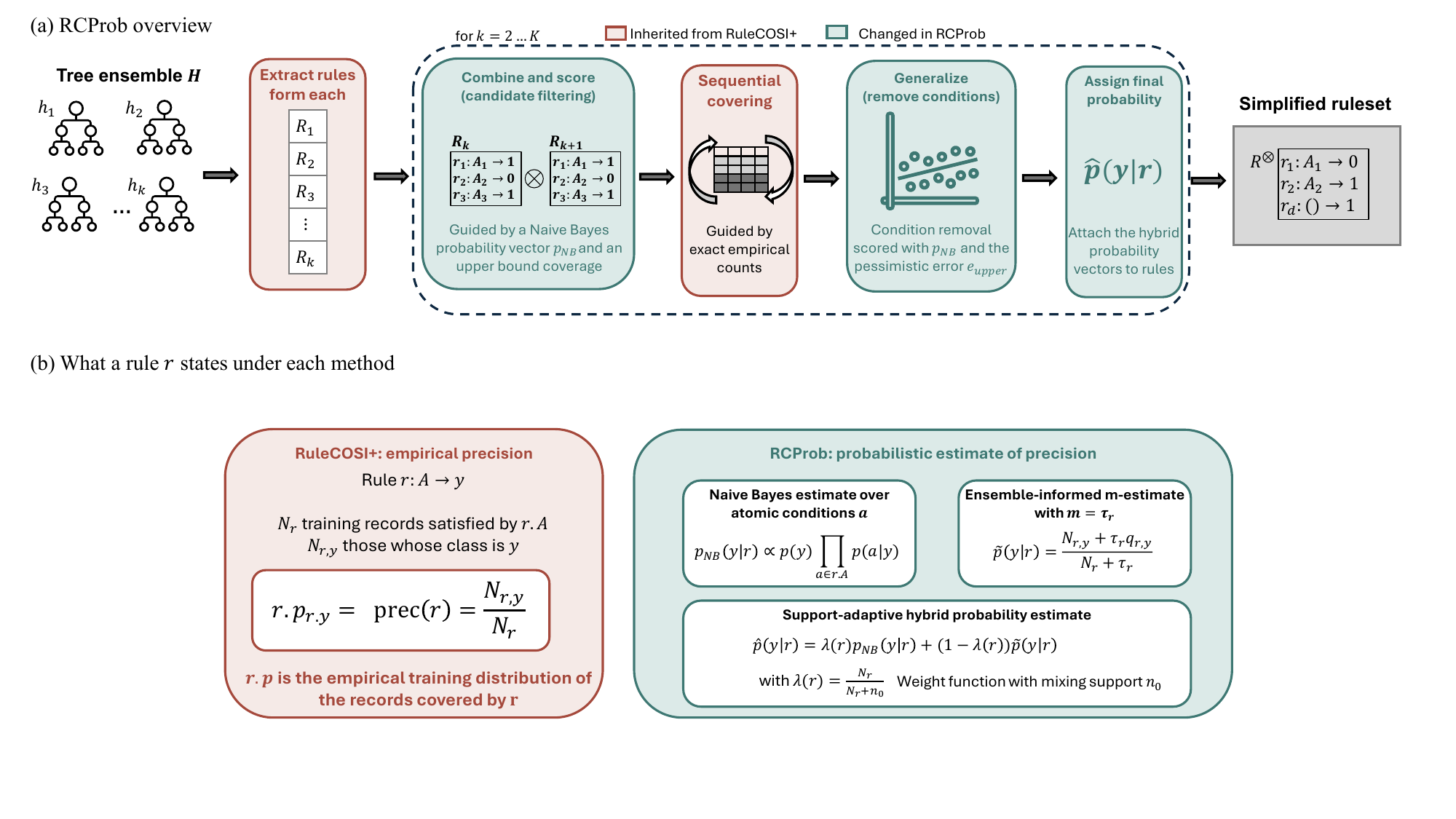}
\caption{Overview of RCProb. (a) The RC+ pipeline, with the stages
RCProb changes highlighted. (b) The two probability estimators for a single
rule: RC+ states the training precision, RCProb a support-adaptive
mixture of the Naive Bayes and m-estimate probabilities.}
\label{fig:rcprob-overview}
\end{figure*}

\subsection{Overview}
\label{sec:m-overview}

Figure~\ref{fig:rcprob-overview} places the two estimates within the RC+
pipeline, and Algorithm~\ref{alg:rcprob} summarizes the implementation used
in the experiments. Each tree is converted into a raw ruleset by a map
$\psi$ that sends every root-to-leaf path to a rule. Exact training support
is computed for the raw rules and the hybrid probability vector described in
Section~\ref{sec:m-mixture} is attached to them. The first ruleset is
simplified, after which the remaining tree rulesets are processed in
sequence.

For a merge candidate, RCProb first computes a Naive Bayes probability
vector and a cheap upper bound on intersection coverage. Candidates failing
these early checks are discarded without computing exact counts. Candidates
that survive are scored with the Naive Bayes probability and an approximate
coverage estimate and then enter the RC+ simplification procedure.
Sequential covering retains empirical counting. During generalization,
condition removals are evaluated with the approximate Naive Bayes
statistics. After generalization, support and class counts are recomputed
and the hybrid probability vector is attached to each surviving rule. The
better of the new and previous rulesets is retained, and a default rule is
added after the combination loop.

\begin{algorithm}[htbp]
\footnotesize
\caption{RCProb}
\label{alg:rcprob}
\begin{algorithmic}[1]
\Require ensemble $H=\{h_1,\ldots,h_K\}$, training set $D$, precision threshold $\eprec$, coverage threshold $\ecov$, confidence level $c$ for the pessimistic
error estimate,
smoothing strength $\eta$, prior strength $\tau$, mixing support $n_0$
\Ensure simplified ruleset $R^{\otimes}$
\State Precompute smoothed class priors and atomic class-conditional
likelihoods using $D$ and $\eta$
\State Extract raw rulesets $R_k\gets\psi(h_k)$ and compute their
support and hybrid probability vectors
\State $R^{\otimes}\gets\textsc{SimplifyRuleset}(R_1)$
\For{$k=2$ to $K$}
  \State $R'\gets\emptyset$
  \For{each pair $(r_i,r_j)\in R^{\otimes}\times R_k$}
    \State Merge and simplify the antecedents
    \State Compute $p_{\mathrm{NB}}(y\mid r)$ and the coverage upper bound
    \If{the early coverage or precision check fails}
      \State discard the pair
    \Else
      \State Estimate merge coverage and score the candidate with
      $p_{\mathrm{NB}}(y\mid r)$
      \If{the approximate thresholds are satisfied}
        \State add the constructed merge candidate to $R'$
      \EndIf
    \EndIf
  \EndFor
  \State $R'\gets\textsc{SequentialCoveringPruning}(R',D)$ using
  empirical statistics
  \State $R'\gets\textsc{GeneralizeRuleset}(R')$ using $p_{\mathrm{NB}}(y\mid r)$ during condition removal
  \State Recompute $N_{r}$ and $N_{r,y}$ for $R'$ and attach the
  hybrid probability vectors $\widehat{p}(y\mid r)$
  \State $R^{\otimes}\gets$ the better of $R^{\otimes}$ and $R'$
\EndFor
\State Append the default rule for records not covered by $R^{\otimes}$
\State \Return $R^{\otimes}$
\end{algorithmic}
\end{algorithm}

\subsection{Naive Bayes estimate over atomic conditions}
\label{sec:m-nb}

Let the antecedent of rule $r$ contain atomic conditions $a\in r.A$. RCProb uses these conditions as independent evidence contributors. Writing
$p(y)$ for the class prior and $p(a\mid y)$ for the likelihood of an atomic
condition, both smoothed as defined below, the estimate for each class
$y\in\mathcal{Y}$ is
\begin{equation}
\label{eq:nb}
p_{\mathrm{NB}}(y\mid r) \propto
p(y)\prod_{a\in r.A}p(a\mid y).
\end{equation}
The computation is performed in log space,
\begin{equation}
\label{eq:nblog}
\log p_{\mathrm{NB}}(y\mid r)
=\log p(y)+\sum_{a\in r.A}\log p(a\mid y)+\mathrm{const.}
\end{equation}
The constant is the same for every class and cancels when the vector is
normalized over $\mathcal{Y}$. The sum over the atomic conditions is the
accumulated log evidence reused at merge time in
Equation~\ref{eq:evidence}.

The prior and the likelihoods are smoothed additively with strength $\eta$.
With $N_y$ training records from class $y$, the smoothed class prior is
\begin{equation}
\label{eq:prior}
p(y)=\frac{N_y+\eta}{n+C\eta}.
\end{equation}
For an atomic condition $a$, let $N_{a,y}$ be the number of class-$y$
training records satisfying that condition. Its smoothed class-conditional
likelihood is
\begin{equation}
\label{eq:likelihood}
p(a\mid y)=\frac{N_{a,y}+\eta}{N_y+2\eta}.
\end{equation}
The value $\eta=1$ gives Laplace smoothing and is fixed in the main
experiments. Equation~\ref{eq:prior} is the general form of the Laplace
estimate, in which the number of classes appears in the denominator
\citep{furnkranzFoundationsRuleLearning2012}. Naive Bayes probabilities are not assumed to be calibrated by
construction; dependence among conditions can produce overconfident
estimates \citep{silva2023classifier,kullBetaCalibrationWellfounded2017}.
RCProb uses this posterior as an inexpensive evidence-composition estimate
during search and regularizes the probability attached to the final rule
through the hybrid estimator in Section~\ref{sec:m-mixture}. The class
priors and atomic likelihoods depend on $D$ and the conditions extracted
from the ensemble, not on a particular merged rule. They are therefore
computed once and reused during the search.

\subsection{Ensemble-informed m-estimate}
\label{sec:m-leaf}

The second estimate is an m-estimate of the rule probability
\citep{furnkranzFoundationsRuleLearning2012}, with the prior taken from the
ensemble. An m-estimate corrects the empirical proportion towards a prior
probability, and the correction is stronger for rules of low coverage. In its
standard form the prior is the class distribution of the training set, so
every rule is corrected towards the same point. RCProb instead corrects
each rule towards a local class distribution obtained from the source
ensemble for the region of that rule. Let
$\mathbf{q}_r=(q_{r,y})_{y\in\mathcal{Y}}$ denote that distribution. For a
rule extracted from a bagging ensemble, $\mathbf{q}_r$ is the class
distribution of the source leaf. For a boosting ensemble the leaf stores a
score, and $\mathbf{q}_r$ is obtained by mapping the region's score vector
through the logistic function for two classes and the softmax for more.
When two rules are merged, their local distributions are combined
elementwise and normalized,
\begin{equation}
\label{eq:localprior}
\mathbf{q}_{r_1\cap r_2}
=\frac{\mathbf{q}_{r_1}\odot\mathbf{q}_{r_2}}
{\left\lVert\mathbf{q}_{r_1}\odot\mathbf{q}_{r_2}\right\rVert_1}.
\end{equation}
The product treats the two source leaves as independent evidence about the
class. For a boosting ensemble it is equivalent to adding the scores of the
two trees, which is how such an ensemble aggregates. For a bagging ensemble
it is sharper than the average that $\phi$ applies, so a merged rule is
given a prior more concentrated than the ensemble's own estimate for the
region.

The prior strength $\tau$ sets how many prior observations a rule is given.
For a rule with exact support count $N_r$ and class count $N_{r,y}$, define
$\tau_r=\min(\tau,N_r)$. The ensemble-informed m-estimate is
\begin{equation}
\label{eq:leaf}
\widetilde{p}(y\mid r)
=\frac{N_{r,y}+\tau_r q_{r,y}}{N_r+\tau_r},
\qquad N_r>0.
\end{equation}
This is the m-estimate with $m=\tau_r$ and prior $q_{r,y}$, so $\tau$ sets the
number of prior observations attributed to a rule. The cap
$\tau_r=\min(\tau,N_r)$ departs from the standard form, in which $m$ is
constant. It keeps the prior from carrying more than half of the weight when a
rule is supported by fewer than $\tau$ records.

\subsection{Support-adaptive hybrid probability}
\label{sec:m-mixture}

The two probability estimates are combined using the support count $N_r$. The mixing support $n_0$ is the support count at which
the two receive equal weight. The Naive Bayes weight is
\begin{equation}
\label{eq:lambda}
\lambda(r)=\frac{N_r}{N_r+n_0}
=\frac{n\,\mathrm{cov}(r)}{n\,\mathrm{cov}(r)+n_0}.
\end{equation}
The probability vector attached to a rule after exact recomputation is
\begin{equation}
\label{eq:hybrid}
\widehat{p}(y\mid r)
=\lambda(r)p_{\mathrm{NB}}(y\mid r)
+\bigl(1-\lambda(r)\bigr)\widetilde{p}(y\mid r).
\end{equation}
Sparse rules receive more weight from the ensemble-informed estimate. As support grows, the Naive Bayes estimate receives progressively
more weight. The weight is a fixed function of rule support and is not fitted
to the data.

\subsection{Approximate merge scoring and evidence reuse}
\label{sec:m-merge}

Candidate generation is an expensive inner loop of RC+. RCProb avoids
recomputing empirical class counts for every pair before the early filtering
decision. For class $y$, define the accumulated log evidence
\begin{equation}
\label{eq:evidence}
E_y(r)=\sum_{a\in r.A}\log p(a\mid y).
\end{equation}
For two rules merged by conjunction, duplicated conditions are removed by
subtracting the evidence of the intersection,
\begin{equation}
\label{eq:merge}
E_y(r_1\cap r_2)
=E_y(r_1)+E_y(r_2)-E_y(r_1.A\cap r_2.A).
\end{equation}
The resulting evidence in Equation~\ref{eq:merge} gives
$p_{\mathrm{NB}}(y\mid r)$ without scanning $D$. The early coverage check
uses $\min(N_{r_1},N_{r_2})/n$ as an upper bound on the coverage of the
intersection. For pairs that pass this check, merge scoring uses the
independence approximation
\begin{equation}
\label{eq:mergecov}
\widetilde{N}_{r_1\cap r_2}
=\frac{N_{r_1}N_{r_2}}{n}
\end{equation}
and the Naive Bayes probability for the precision check. Because exact
sequential covering is retained, a bitset for a surviving intersection is
materialized for later use, but the merge stage avoids the exact population
counts used by RC+.

\subsection{Integration with RC+ simplification}
\label{sec:m-integration}

Sequential covering pruning keeps the RC+ empirical procedure. At each
iteration, rule statistics are recomputed on the currently uncovered
training records, the rules are ordered, and the first rule satisfying the
coverage threshold $\ecov$ and the precision threshold $\eprec$ is retained.

Generalization removes conditions while the pessimistic error estimate
improves. Following RC+, the statistical correction is
\begin{equation}
\label{eq:pessimistic}
e_{\mathrm{upper}}(N_r,e_r,c)=
\frac{e_r+\frac{z_{c/2}^{2}}{2N_r}
+z_{c/2}\sqrt{\frac{e_r(1-e_r)}{N_r}
+\frac{z_{c/2}^{2}}{4N_r^{2}}}}
{1+\frac{z_{c/2}^{2}}{N_r}},
\end{equation}
where $e_r$ is the estimated error for the consequent class and $z_{c/2}$ is
the corresponding standard-normal value. In the RCProb configuration used in
this study, candidate condition removals are evaluated with the approximate
coverage and Naive Bayes class probabilities from Section~\ref{sec:m-nb}.
Once the generalization search finishes, support and class counts are
recomputed and Equation~\ref{eq:hybrid} is used for the probability vector
attached to the surviving rule.

\subsection{Complexity analysis}
\label{sec:m-complexity}

We follow the worst-case analysis used for RC+
\citep{obregonRuleCOSIRuleExtraction2023}. A tree of depth $d$ contains at
most $2^d$ terminal rules. Pairwise combination therefore gives a $4^d$
term, while the condition-removal search in generalization contributes the
dominant $8^d$ term. RC+ computes empirical rule heuristics from the
training data, giving the published bound $O(Kn8^d)$.

RCProb precomputes the atomic tables once at $O(nK2^d)$. Its approximate
probability scoring removes the factor of $n$ from the condition-removal
search, giving $O(C8^d)$ for that part. The implementation still retains
data-dependent work for constructed merge candidates and exact sequential
covering, both bounded by an $n4^d$ term. The resulting worst-case bound is
\begin{equation}
\label{eq:complexity}
O\!\left(K\left(n4^d+C8^d\right)\right),
\end{equation}
where the one-time $nK2^d$ precomputation is absorbed by the $nK4^d$ term.
The bound characterizes the computational structure of the two procedures. 
Changing the probability estimator also changes which merge candidates pass the early
checks and therefore changes the subsequent search trajectory.

% The implementation records \texttt{n\_combinations\_} after a pair has
% passed the earliest zero-coverage and cached-bad-combination checks and
% reaches rule construction. We refer to this quantity as the number of
% \emph{constructed merge candidates}. It is neither the full Cartesian
% product of rule pairs considered by the loops nor the number of rules that
% survive all later thresholds and simplification.

%% file: sections/05_experiments.tex
\section{Experimental design}
\label{sec:experiments}

The experiments address three questions. RQ1 examines whether the
probability values attached to extracted RC+ rules remain reliable on
held-out data. RQ2 evaluates how RCProb changes probabilistic quality,
ruleset compactness, classification performance, fidelity, and computational
cost relative to RC+. RQ3 compares the native RCProb probabilities with RC+
followed by a standard post-hoc calibration stage. Table~\ref{tab:rqmap}
maps each question to its main evidence.

\begin{table*}[htbp]
\centering
\caption{Research questions and the evidence used to answer them.}
\label{tab:rqmap}
\small
\begin{tabular}{p{0.06\textwidth}p{0.24\textwidth}p{0.26\textwidth}p{0.23\textwidth}}
\toprule
 & Question & Analysis & Main evidence \\
\midrule
RQ1 & Are the probability values attached to extracted RC+ rules
reliable on unseen data? &
Held-out probabilistic evaluation of RC+ and descriptive examination
of the probability stated by individual decision-list rules &
Log-loss, confidence-ECE, classwise-ECE and Brier score; per-rule subgroup
calibration gaps and prevalence of near-certain rules \\
\addlinespace
RQ2 & How does RCProb affect probabilistic quality, ruleset compactness,
classification performance, fidelity, and computational cost relative to
RC+? &
Paired RC+--RCProb comparison using the same trained ensemble in every
outer fold &
Probabilistic scores, macro-F1, fidelity, ruleset size and antecedent
measures, and simplification time \\
\addlinespace
RQ3 & How does RCProb compare with RC+ after a conventional post-hoc
calibration stage? &
Separate controlled experiment with dedicated training, calibration and
test partitions &
Raw RCProb versus temperature-scaled RC+ on log-loss, Brier score,
confidence-ECE and classwise-ECE \\
\bottomrule
\end{tabular}
\end{table*}

\subsection{Datasets}
\label{sec:ex-datasets}

The primary evaluation uses \NDatasets{} publicly available benchmark
classification datasets, summarized in Table~\ref{tab:datasets}. Eighteen
are binary and five are multiclass with between three and nine classes. They
range from 90 to 17{,}898 records and from 4 to 57 attributes, and their
imbalance ratios span 1.02 to 28.50. During dataset preparation, rows with
missing values are removed and a uniform minimum class-support rule removes
classes represented by fewer than 10 observations before label encoding.
This rule affects \textit{ecoli} and \textit{yeast};
Table~\ref{tab:datasets} reports the exact post-preprocessed records used
in the experiments. The counts can therefore differ slightly from commonly
cited repository totals or from other retrieved versions of the same
dataset. Stratified partitioning is used throughout to preserve the class
distribution in each split.

\input{tables/t1_datasets}

\subsection{Base ensembles and compared methods}
\label{sec:ex-methods}

Two base ensembles are used: random forest \citep{Breiman2001} for an
independently constructed ensemble and gradient boosting machine
\citep{Friedman2001} for a dependent ensemble. For every outer fold, RC+ and
RCProb receive an ensemble fitted with the same selected hyperparameters,
training data and random seed. 

The primary comparison is between RC+
\citep{obregonRuleCOSIRuleExtraction2023} and RCProb under decision-list
inference. This is the setting in which the first matching rule supplies the
probability vector for an instance and therefore provides the clearest link
between a rule's stated probability and the prediction delivered to the
user. Aggregate inference, in which all matching non-default rule
probability vectors are summed and normalized (Equation~\ref{eq:aggregate}),
is retained as a sensitivity analysis in Table~S10.

Three additional methods provide context. inTrees \citep{Deng2019} and DFT
\citep{haraMakingTreeEnsembles2018} simplify the same fitted base ensembles.
DT is a single decision tree fitted on the outer training partition and
tuned over depth and cost-complexity pruning, with each leaf stating the
empirical class frequencies of the training records it captures. It does not
read the ensemble, so it serves as an interpretable reference for what a
comparable ruleset achieves without extraction, and fidelity does not apply
to it. All comparison methods use the
same outer partitions as RC+ and RCProb. 

\subsection{Cross-validation and hyperparameter optimization}
\label{sec:ex-protocol}

The primary benchmark uses stratified $3\times5$ cross-validation, that is,
five-fold cross-validation repeated three times. This produces \NFolds{}
held-out evaluations for each dataset and base ensemble. Within each outer
training partition, ensemble hyperparameters are selected first by
three-fold cross-validation. The outer test fold is used only for final 
evaluation. The search varies the number of trees
$n_{\mathrm{trees}}\in\{50,100,250\}$ and the maximum depth
$d\in\{3,4,5,6\}$.

Table~\ref{tab:hpo} gives the search spaces for all models. RC+ and RCProb grids are
method-specific because the smoothed RCProb probabilities change the useful
operating ranges. RCProb omits the near-certain 0.95 precision threshold and
retains the lower 0.001 coverage threshold. The pessimistic-error grid is shared. The three additional
RCProb parameters are fixed globally: $\eta=1$, $\tau=5$, and $n_0=50$. Here
$\eta=1$ is Laplace smoothing. The support-mixing and prior-strength choices
were examined in targeted sensitivity analyses, reported in Table~S13.
Those reduced studies show a nearly flat low-$\tau$
region and no improvement from increasing $n_0$ beyond 50. They assess the
stability of these settings, and the main nested-CV benchmark remains the
source of the reported performance.

\input{tables/t2_hpo}

\subsection{Evaluation measures}
\label{sec:measures}

The probability estimates are evaluated with complementary standard measures
because calibration error and overall probabilistic quality are not the same
property. In particular, a predictor can obtain a small calibration error by
issuing uninformative probabilities. Proper scoring rules should therefore
accompany calibration-specific measures
\citep{silva2023classifier,kullTemperatureScalingObtaining2019,
gneitingStrictlyProperScoring2007}.

\subsubsection{Probabilistic quality}

Let $\mathbf{p}_i=(p_{i1},\ldots,p_{iC})$ be the probability vector
predicted for test instance $i$, over the $N_{\mathrm{test}}$ held-out
records of a fold. The logarithmic loss is
\begin{equation}
\label{eq:logloss}
\mathrm{LogLoss}
=-\frac{1}{N_{\mathrm{test}}}\sum_{i=1}^{N_{\mathrm{test}}}\log p_{i,y_i}.
\end{equation}
Lower values are better and zero is optimal. The logarithmic loss is a
strictly proper scoring rule and assigns a large penalty when a model gives
very low probability to the class that actually occurs
\citep{silva2023classifier,gneitingStrictlyProperScoring2007}. This
property is directly relevant to the near-certain but incorrect predictions
motivating RCProb. For computation, the probability assigned to the true
class is clipped below at $10^{-15}$ before taking the logarithm, avoiding
infinite values when a method assigns zero probability to the observed
class. The Brier score is reported alongside log-loss as a bounded proper
scoring rule that does not depend on this numerical clipping.

The multiclass Brier score is
\begin{equation}
\label{eq:brier}
\mathrm{BS}
=\frac{1}{N_{\mathrm{test}}}\sum_{i=1}^{N_{\mathrm{test}}}\sum_{y\in\mathcal{Y}}
\left(p_{i,y}-\mathbb{I}[y_i=y]\right)^2.
\end{equation}
It is also a proper scoring rule, but its penalty is quadratic, not
logarithmic. We use the full probability-vector convention in
Equation~\ref{eq:brier}; for a binary problem it is twice the value obtained
when the Brier score is written only for the positive-class probability. The
full-vector convention follows the calibration literature
\citep{silva2023classifier}. The Brier score is computed at the instance
level from the probability vector returned by the ruleset.

\subsubsection{Calibration error}

Confidence-ECE follows the usual top-class formulation
\citep{guoCalibrationModernNeural2017,silva2023classifier}. Let
$\mathrm{conf}_i=\max_y p_{i,y}$ and let $z_i$ indicate whether the class
with maximum probability is correct. Predictions are partitioned into $B=10$
equal-width confidence bins. For bin $B_b$, let $\mathrm{conf}(B_b)$ and
$\mathrm{acc}(B_b)$ be its mean confidence and accuracy. Then
\begin{equation}
\label{eq:ece}
\mathrm{ECE}
=\sum_{b=1}^{B}\frac{|B_b|}{N_{\mathrm{test}}}
\left|\mathrm{acc}(B_b)-\mathrm{conf}(B_b)\right|.
\end{equation}
ECE directly targets the reliability of the top-class confidence, but it
depends on the binning scheme and does not evaluate the remaining class
probabilities.

For multiclass predictions, classwise-ECE evaluates every class probability,
not only the maximum probability
\citep{kullTemperatureScalingObtaining2019}. For class $y$, the values
$p_{i,y}$ are partitioned into the same number of equal-width bins and
compared with the observed frequency of class $y$ in each bin. The result is
averaged across classes,
\begin{equation}
\label{eq:cwece}
\mathrm{cwECE}
=\frac{1}{C}\sum_{y=1}^{C}\sum_{b=1}^{B}
\frac{|B_{b,y}|}{N_{\mathrm{test}}}
\left|\overline{p}_{b,y}-\overline{o}_{b,y}\right|.
\end{equation}
Confidence-ECE is the more direct measure for the confidence
value shown with a rule; classwise-ECE is retained as a multiclass
robustness check.

Under decision-list inference, the model-level probability vector in all
four measures is exactly the probability vector supplied by the first rule
that fires. Thus the standard probabilistic scores evaluate the probability
statements delivered by the rules without requiring a nonstandard rule-level
scoring rule.

\subsubsection{Rule-level descriptive calibration}

Each decision-list rule defines a subgroup of held-out observations. For
each such subgroup we compute the signed difference $g_r$ between the stated
probability and the held-out precision, defined in
Section~\ref{sec:pre-measures}. We report the distribution of these gaps by
held-out rule support and the fraction of non-default rules with a stated
probability of at least 0.999. Both are diagnostics of the failure mode and
sit outside the inferential metric family.

\subsubsection{Classification, fidelity, and ruleset structure}

Macro-F1 is used to evaluate hard classification because it assigns equal
weight to every class. Fidelity is the fraction of outer-test observations
for which the simplified ruleset and its source ensemble make the same hard
prediction. Ruleset structure is measured by the number of rules, the total
number of antecedent conditions, the number of distinct antecedent
conditions, the mean number of conditions per rule, and the antecedent
uniqueness ratio \textsc{uniq} introduced with RC+
\citep{obregonRuleCOSIRuleExtraction2023}. Let $A$ be the set of
conditions appearing in a ruleset $R$ and $A^*$ the corresponding
multiset, which retains one element per occurrence. The uniqueness ratio
is
\begin{equation}
\label{eq:uniq}
\textsc{uniq}(R)=\frac{|A|}{|A^*|},
\end{equation}
which lies in $(0,1]$ and is higher for a ruleset that repeats fewer
conditions across its rules. The rule count includes the default rule, whose antecedent
is empty.

\subsection{Controlled post-hoc calibration experiment}
\label{sec:ex-posthoc}

The primary cross-validation benchmark did not reserve a calibration set, so
post-hoc calibration is evaluated in a separate controlled experiment and is
never fitted to the outer test folds. The experiment uses six datasets
selected before model fitting by size and class representation:
\textit{htru2}, \textit{maintenance}, \textit{occupancy},
\textit{pageblocks0}, \textit{spambase}, and \textit{vehicle}. For each RF
and GBM experiment, five repeated stratified splits allocate 60\% of the
data to model development, 20\% to calibration and 20\% to final testing.
All ensemble and simplifier HPO is contained within the 60\% development
partition.

Temperature scaling is the principal post-hoc baseline. It learns a single
parameter from held-out calibration predictions and preserves the predicted
class \citep{guoCalibrationModernNeural2017}. The calibration partition is
split internally into three folds, following the non-neural calibration
protocol of \citet{kullTemperatureScalingObtaining2019}, and the three
calibration-map predictions are averaged on the test set. Raw RCProb is
compared descriptively with temperature-scaled RC+ using log-loss, Brier
score, confidence-ECE and classwise-ECE. The same calibration procedure is
also applied to RCProb as an exploratory measure of remaining calibration
headroom.

A more flexible L2-regularized Dirichlet calibrator
\citep{kullTemperatureScalingObtaining2019} is evaluated as a sensitivity
analysis. Because it can change the predicted class and is more demanding of
the calibration sample, its detailed results, together with the
post-calibrated RCProb results, are placed in Table~S8.
The controlled calibration experiment is reported descriptively, and its six
datasets support comparison of the calibration strategies but not a separate
claim of statistical difference.

\subsection{Statistical analysis}
\label{sec:ex-stats}

RF and GBM are analyzed separately because they represent different ensemble
construction frameworks. The independent unit is the dataset. For each
method, base ensemble, and dataset, the 15 held-out values from three
repetitions and five outer folds are averaged before inference. 

The primary inferential comparison is RCProb versus RC+ under decision-list
inference. Within each base ensemble, the resulting \NDatasets{} paired
dataset means are compared by a two-sided Wilcoxon signed-rank test
\citep{Demsar2006}. Twelve outcomes form one multiplicity family: log-loss,
Brier score, confidence-ECE, classwise-ECE, macro-F1, fidelity, number of
rules, total antecedent conditions, distinct antecedent conditions,
conditions per rule, \textsc{uniq}, and simplification fit time. Holm's
step-down procedure controls the family-wise error rate at 0.05 separately
for RF and GBM. The unadjusted and Holm-adjusted $p$-values and the median
paired relative changes are reported in Table~\ref{tab:primary-tests}.

A secondary five-method benchmark compares RCProb, RC+, DT, inTrees, and
DFT separately for RF and GBM. For each outcome, a Friedman test first
examines whether the five methods have the same distribution of ranks across
datasets. When the omnibus test is significant at 0.05, RCProb is compared
with each of the four alternatives using paired Wilcoxon signed-rank tests,
with Holm correction across those four post-hoc comparisons. The five-method
test covers the interpretable models only. The source ensemble is the model
being simplified, so it appears throughout as a descriptive reference. The
complete post-hoc matrices are reported in Tables~S4 and~S5, and the average
ranks in Tables~S6 and~S7. When the primary RCProb--RC+ family and the
secondary benchmark lead to different adjusted decisions, the predefined
primary family guides the RCProb--RC+ claim.

The per-rule subgroup diagnostic, search replay, parameter sensitivity,
aggregate-inference robustness, and controlled post-hoc calibration
experiment are secondary analyses. They are reported separately from these
inferential families.

%% file: tables/t1_datasets.tex
\begin{table}[htbp]
\centering
\caption{Characteristics of the 23 benchmark datasets. IR is the ratio of the largest to the smallest class. For multiclass datasets, the listed classes are the smallest and largest classes.}
\label{tab:datasets}
\small
\begin{tabular}{llcrrr}
\toprule
Dataset & Class (minor; major) & \#cls & IR & \#inst. & \#atts. \\
\midrule
\multicolumn{6}{l}{\textit{Binary}} \\
\midrule
cryotherapy & (0; 1) & 2 & 1.14 & 90 & 6 \\
divorce & (1; 0) & 2 & 1.02 & 170 & 54 \\
heart & (0; 1) & 2 & 1.25 & 270 & 13 \\
ionosphere & (0; 1) & 2 & 1.79 & 351 & 34 \\
bands & (0; 1) & 2 & 1.70 & 365 & 19 \\
credit & (1; 0) & 2 & 11.11 & 666 & 51 \\
wisconsin & (0; 1) & 2 & 1.86 & 683 & 9 \\
australian & (1; 0) & 2 & 1.25 & 690 & 14 \\
pima & (0; 1) & 2 & 1.87 & 768 & 8 \\
mammographic & (1; 0) & 2 & 1.06 & 830 & 5 \\
biodeg & (0; 1) & 2 & 1.96 & 1,055 & 41 \\
banknote & (1; 0) & 2 & 1.25 & 1,372 & 4 \\
cardiotoc & (0; 1) & 2 & 3.51 & 2,126 & 21 \\
spambase & (1; 0) & 2 & 1.54 & 4,601 & 57 \\
pageblocks0 & (0; 1) & 2 & 8.79 & 5,472 & 10 \\
occupancy & (0; 1) & 2 & 3.71 & 8,143 & 5 \\
maintenance & (1; 0) & 2 & 28.50 & 10,000 & 8 \\
htru2 & (1; 0) & 2 & 9.92 & 17,898 & 8 \\
\midrule
\multicolumn{6}{l}{\textit{Multiclass}} \\
\midrule
glass & (4; 1) & 6 & 8.44 & 214 & 9 \\
new-thyroid & (2; 0) & 3 & 5.00 & 215 & 5 \\
ecoli & (3; 0) & 5 & 7.15 & 327 & 7 \\
vehicle & (3; 2) & 4 & 1.09 & 845 & 18 \\
yeast & (7; 0) & 9 & 23.15 & 1,479 & 8 \\
\bottomrule
\end{tabular}
\end{table}

%% file: tables/t2_hpo.tex
\begin{table*}[htbp]
\centering
\caption{Hyperparameter search spaces. RC+ and RCProb use method-specific threshold grids but equal 18-configuration search budgets. The RCProb parameters $\eta=1$, $\tau=5$, and $n_0=50$ are fixed globally and examined in targeted sensitivity analyses.}
\label{tab:hpo}
\small
\resizebox{\textwidth}{!}{%
\begin{tabular}{p{0.15\textwidth}p{0.70\textwidth}r}
\toprule
Method & Search space & Configurations \\
\midrule
Base ensemble & $n_{\mathrm{trees}}\in\{50,100,250\}$, $d\in\{3,4,5,6\}$ & 12 \\
\addlinespace
RCProb & $\eprec\in\{0.5,0.75\}$, $\ecov\in\{0,0.001,0.01\}$, $c\in\{0.10,0.25,0.40\}$ & 18 \\
RC+ & $\eprec\in\{0.5,0.75,0.95\}$, $\ecov\in\{0,0.01\}$, $c\in\{0.10,0.25,0.40\}$ & 18 \\
\addlinespace
inTrees & $\mathrm{maxdecay}\in\{0.05,0.01\}$, $\mathrm{minfreq}\in\{0.1,0.01\}$ & 4 \\
DefragTrees & $\mathrm{maxitr}\in\{50,100\}$, $\mathrm{maxrules}\in\{4,8,16\}$ & 6 \\
DT & $d\in\{\mathrm{none},2,3,4,6\}$, $\alpha_{\mathrm{ccp}}\in\{0,0.005,0.01\}$ & 15 \\
\bottomrule
\end{tabular}%
}
\end{table*}

%% file: sections/06_results.tex
\section{Results}
\label{sec:results}

The results are reported separately for RF and GBM because they represent
independent and dependent ensemble construction frameworks, respectively.
The primary statistical comparison is RCProb versus RC+ within each base
ensemble. The source ensemble and the three additional interpretable methods
provide context. Table~\ref{tab:primary-tests} reports the complete primary
comparison, and every Holm-adjusted $p$-value quoted in this section is taken
from it. Full per-dataset results (Tables~S2 and~S3) and the secondary
five-method tests (Tables~S4 and~S5) are reported in the Supplementary
Material.

\subsection{Random forest results}
\label{sec:res-rf}

Table~\ref{tab:rf-main} summarizes the RF experiments. RCProb reduces mean
log-loss from 1.9003 for RC+ to 0.4428. The median paired relative reduction
is 71.9\%, and the difference remains significant after Holm correction
($p_{\mathrm{Holm}}<0.0001$). Confidence-ECE decreases from 0.0895 to
0.0623, with a median reduction of 25.4\% ($p_{\mathrm{Holm}}=0.0362$).
Classwise-ECE also decreases, but the adjusted $p$-value is 0.0743. The
Brier-score difference is not significant after correction.

Macro-F1 is 0.8339 for RC+ and 0.8278 for RCProb, while fidelity is 0.9088
and 0.9161, respectively. Neither difference is statistically distinguished
under the primary 12-outcome family. These results do not establish
equivalence, but the observed hard-classification differences are small
relative to the log-loss change.

RCProb also returns a smaller RF ruleset. The mean rule count decreases from
35.82 to 20.00, with a median paired reduction of 38.7\%
($p_{\mathrm{Holm}}<0.0001$). Total antecedent conditions decrease by 20.1\%
($p_{\mathrm{Holm}}=0.0191$), and distinct antecedents decrease by 34.8\%
($p_{\mathrm{Holm}}<0.0001$). The retained rules are longer: conditions per
rule increase by 17.3\% ($p_{\mathrm{Holm}}=0.0102$). The \textsc{uniq}
ratio is also lower for RCProb ($p_{\mathrm{Holm}}=0.0057$). Simplification
time is reported in Section~\ref{sec:res-cost}.

The comparison methods show different tradeoffs. RCProb has the lowest mean
log-loss of the five interpretable models, and the secondary benchmark
distinguishes it from every comparator on that measure. Against the decision
tree the two are indistinguishable in Brier score, macro-F1, fidelity, and
rule count, so the log-loss difference does not come with a classification or
size penalty; the tree is far cheaper to fit. RCProb has significantly better
log-loss, Brier score, macro-F1, and fidelity than inTrees and DFT. inTrees
and DFT use fewer antecedents, and DFT produces the smallest rulesets and the
lowest numerical ECE values but has a macro-F1 of only 0.5447, two thirds of every
other method in the table. The five-method Friedman tests for RF
confidence-ECE and classwise-ECE are not significant. The source RF retains the best proper-scoring and
classification performance, with mean log-loss 0.2971 and macro-F1 0.8513.

\input{tables/t3_rf_main}

\subsection{Gradient boosting results}
\label{sec:res-gbm}

Table~\ref{tab:gbm-main} reports the GBM results. RCProb reduces mean
log-loss from 1.6947 to 0.4935, with a median paired reduction of 62.5\%
($p_{\mathrm{Holm}}<0.0001$). Confidence-ECE decreases from 0.0812 to 0.0571
and classwise-ECE from 0.0684 to 0.0519. These reductions do not remain
significant after correction ($p_{\mathrm{Holm}}=0.0596$ and 0.1453,
respectively). The Brier-score difference is also not significant.

Macro-F1 decreases from 0.8378 to 0.8287 and fidelity from 0.9200 to 0.9104.
Neither difference is significant under the primary multiplicity family. The
structural reductions are again substantial. Mean rule count decreases from
25.70 to 14.53, a median reduction of 38.5\% ($p_{\mathrm{Holm}}=0.0006$).
Total and distinct antecedents decrease by 27.1\% and 32.7\%, respectively,
with both differences remaining significant. The increase in conditions per
rule and the decrease in \textsc{uniq} do not remain significant after
correction. In the secondary five-method analysis, the RCProb--RC+ GBM
contrasts for fidelity, \textsc{uniq}, and fit time cross the 0.05 threshold
under the comparator-wise correction. As specified in
Section~\ref{sec:ex-stats}, those secondary decisions do not override the
predefined 12-outcome RCProb--RC+ family used for the primary claims.

RCProb again has the lowest mean log-loss of the five interpretable models,
0.4935 against 1.6842 for the decision tree, and the secondary benchmark
distinguishes the two ($p_{\mathrm{Holm}}<0.0001$). RCProb also uses
significantly fewer total antecedent conditions than the tree
($p_{\mathrm{Holm}}=0.0061$), while the rule-count difference falls just
outside the corrected threshold. RCProb also has significantly better
log-loss, Brier score, macro-F1, and fidelity than inTrees and DFT. inTrees
remains more compact in total antecedent count and rule length, while DFT
has the smallest ruleset and the lowest numerical ECE values. As in RF, the
source GBM retains stronger proper-scoring and classification performance,
with mean log-loss 0.4238 and macro-F1 0.8591.

\input{tables/t3_gbm_main}
\input{tables/s9_primary_tests}

\begin{figure*}[htbp]
\centering
\includegraphics[width=\textwidth]{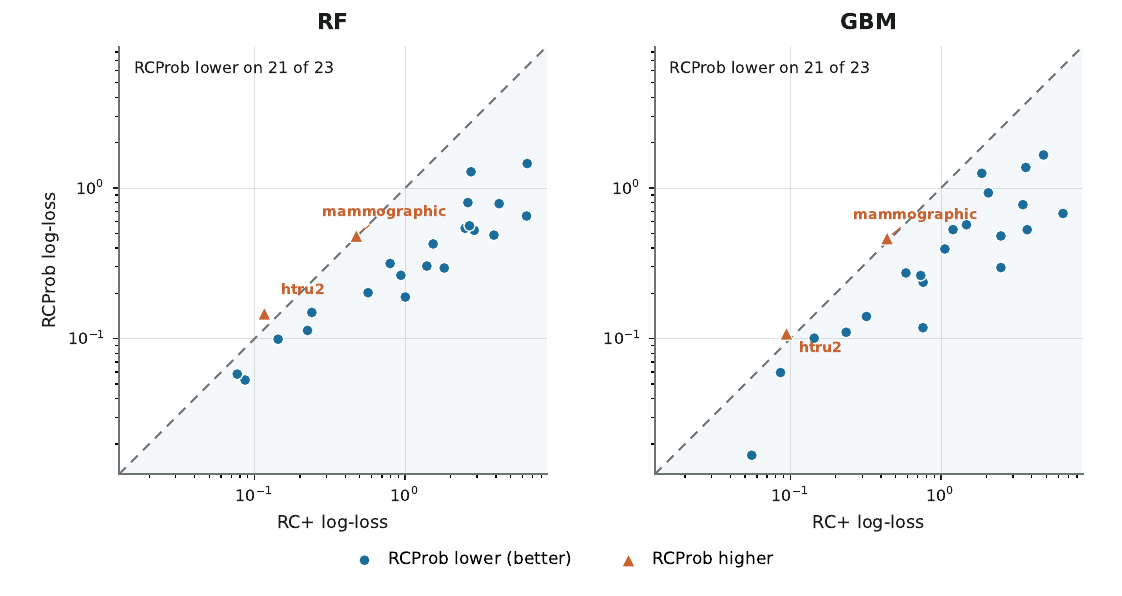}
\caption{Dataset-level RC+ and RCProb log-loss by base ensemble. Each point is
one dataset mean over 15 held-out evaluations, on logarithmic axes. Points
below the dashed equality line favor RCProb. The two triangles mark the
datasets on which RCProb is worse, which are the same two in both base
ensembles. A two-sided Wilcoxon signed-rank test over these 23 paired dataset
means gives $p_{\mathrm{Holm}}<0.0001$ in both base ensembles
(Table~\ref{tab:primary-tests}).}
\label{fig:probability-results}
\end{figure*}

Figure~\ref{fig:probability-results} shows the dataset-level paired log-loss.
The reduction is consistent across both ensemble types, with RCProb lower on
21 of 23 datasets for both RF and GBM, except \textit{htru2} and
\textit{mammographic}. Confidence-ECE is lower
for RCProb on 16 of 23 datasets for RF and 17 of 23 for GBM, although only
the RF difference remains significant under the primary multiplicity
correction. Six of the seven datasets with higher confidence-ECE are common
to the two base ensembles.

\begin{figure*}[htbp]
\centering
\includegraphics[width=\textwidth]{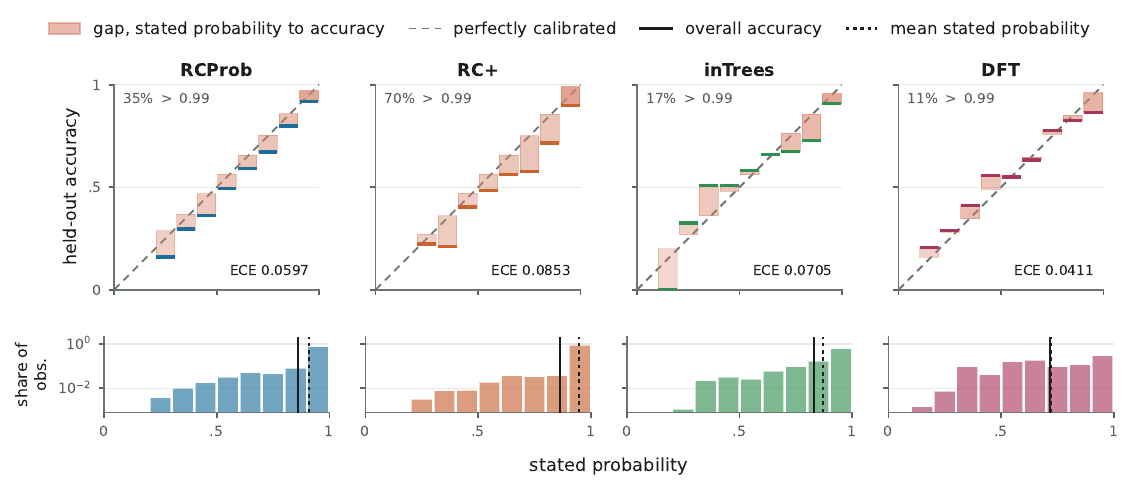}
\caption{Reliability of the stated probability of the predicted class for the
four rule-extraction methods, averaged over the 23 datasets and over both base
ensembles. In the upper row each bar runs from a bin's mean stated probability
to its mean held-out accuracy, so the bar height is the calibration error in
that bin, and the horizontal marker gives the accuracy. The lower row gives
the share of a dataset's held-out observations falling in each bin, on a
logarithmic axis, with the solid and dashed rules marking overall accuracy and
mean stated probability. Bins are equal width, and each dataset carries equal
weight. Each panel is annotated with the share of observations stating more
than 0.99 and with the confidence-ECE of Tables~\ref{tab:rf-main}
and~\ref{tab:gbm-main} averaged over the two base ensembles. }
\label{fig:reliability-methods}
\end{figure*}

Figure~\ref{fig:reliability-methods} shows where in the probability range that
calibration error sits, for RCProb and the three comparators. Every method
states more probability than the held-out data supports, and RC+ by the widest
margin. Most of the difference between RC+ and RCProb lies in how much of the
data sits at the top of the range. RC+ places 70\% of a dataset's held-out
observations above 0.99 against 35\% for RCProb, while the primary tests
detect no macro-F1 difference between them.

\subsection{Computational cost}
\label{sec:res-cost}

Neither base ensemble gives RCProb a simplification-time advantage. Mean fit
time is 13.03~s for RC+ and 13.31~s for RCProb in RF, and 6.76~s and 7.92~s
in GBM (Tables~\ref{tab:rf-main} and~\ref{tab:gbm-main}). The median
dataset-wise relative increases are 25.3\% for RF and 39.8\% for GBM, and
neither difference remains significant under the primary 12-outcome family
($p_{\mathrm{Holm}}=0.3920$ and 0.1369). The dataset-wise medians exceed the
change in the means because each dataset contributes equally to the median
regardless of its absolute runtime.

\subsection{Held-out behavior of individual rule probabilities}
\label{sec:res-rule}

\begin{figure*}[htbp]
\centering
\includegraphics[width=\textwidth]{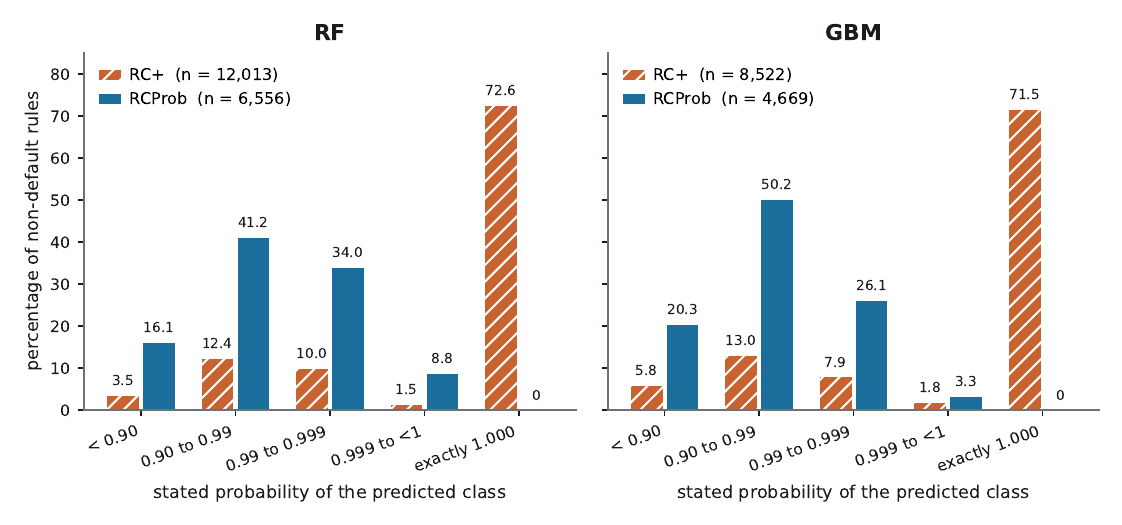}
\caption{Distribution of the stated probability of the predicted class over
all non-default rules, by base ensemble. The bands are unequal because both
methods place most rules above 0.99. The final band isolates rules stating
exactly 1.000.}
\label{fig:confidence-distribution}
\end{figure*}

The rule-level diagnostics in Table~S9 show the same qualitative pattern
for both base ensembles. Among non-default RC+ rules, 72.6\% of RF rules and
71.5\% of GBM rules state a probability of exactly 1.0. No RCProb
non-default rule states a probability of exactly 1.0. At the 0.999
threshold, the proportions are 74.1\% versus 8.8\% for RF and 73.3\% versus
3.3\% for GBM. Figure~\ref{fig:confidence-distribution} shows the complete
distribution. The RCProb rules concentrate between 0.90 and 0.999 in both
base ensembles.

The held-out subgroup analysis also concentrates the largest discrepancies
among sparse rules. For rules assigned 1--9 held-out observations, the mean
dataset-level calibration-in-the-large difference is 0.193 for RC+ and 0.131
for RCProb in RF. The corresponding GBM values are 0.208 and 0.130. The gap
becomes smaller with support and RCProb is not uniformly better in every
support range. These quantities are therefore used descriptively and are
reported in full in Table~S9.

\subsection{Case study: cryotherapy}
\label{sec:res-case}

The \textit{cryotherapy} dataset records the outcome of liquid-nitrogen
cryotherapy for warts in 90 patients
\citep{khozeimehExpertSystemSelecting2017}. Six attributes describe the
patient and the lesion, among them age, the number of months elapsed before
treatment, the wart type, and the surface area in square millimeters. The two
classes indicate whether the treatment succeeded and are close to balanced,
with the exact counts in Table~\ref{tab:datasets}. The data were collected for
an expert system that selects a wart treatment, so the attributes are the ones
recorded before the decision is taken.

Table~\ref{tab:case-cryotherapy} gives one illustrative GBM fold. RC+ and
RCProb retain the same antecedent and consequent, so the example isolates the
probability estimate. The shared rule assigns a patient treated within 8.125
months of onset and aged 50 or under to the class of successful treatments.
RC+ states a class-1 probability of 1.000 for it, whereas RCProb states 0.956.
Six of the seven held-out patients assigned to the rule belong to class 1. The
value 6/7 is an empirical frequency for this fold, not an estimate of the
rule's true probability.

\input{tables/t7_case_cryotherapy}

Tables~\ref{tab:case-rules-rcplus} and~\ref{tab:case-rules-rcprob} give the
complete final decision lists for this fold. All six non-default RC+ rules
state a predicted-class probability of 1.000, whereas the four non-default
RCProb rules range from 0.875 to 0.958. The two methods report the same
training precision for the rules they share, so the difference between
$\mathrm{prec}(r)$ and $r.p_{r.y}$ is the estimator alone. RCProb also
reduces the complete ruleset from seven rules to five, and its second rule
drops the \textit{Area} condition, which widens the covered region from 22
to 33 training records and lowers the precision from 100\% to 90.91\%. The
held-out counts are descriptive and were not used to construct the rulesets.
In this fold the non-default rules cover the whole training partition, so
the default rule has no residual records and falls back to the training
class distribution.

\input{tables/t8_case_rules_rcplus}
\input{tables/t9_case_rules_rcprob}

The spread matters when the ruleset is used to route cases. Under RC+ every
covered patient carries the same stated certainty of 1.000, so no threshold
placed on the stated probability separates them. Under RCProb a referral rule
that asks for a second opinion below 0.90 acts on $r_2$ and on no other rule.
That rule assigns patients treated after 8.125 months and older than 16.5 to
the class of unsuccessful treatments, and it is correct for 8 of the 10
held-out patients for which it fired. This illustrates one use of the stated
probability and is not a clinical recommendation.

\subsection{Controlled post-hoc calibration}
\label{sec:res-posthoc}

Table~\ref{tab:calibration} reports the controlled calibration experiment by
base ensemble. For RF, temperature scaling reduces RC+ mean log-loss from
1.1392 to 0.2904, while raw RCProb obtains 0.2482 without a calibration map.
For GBM, the corresponding values are 0.9674, 0.2691, and 0.2354. Raw RCProb
also has lower Brier score, confidence-ECE, and classwise-ECE than
temperature-scaled RC+ in both base-ensemble experiments. Temperature
scaling substantially improves RC+ log-loss and confidence-ECE, but
classwise-ECE increases from 0.0373 to 0.0522 for RF and from 0.0315 to
0.0412 for GBM. This is consistent with a one-parameter calibration map that
corrects global confidence sharpness without fitting separate class-specific
reliability patterns. Temperature scaling preserves the RC+ hard
predictions, so the RC+ accuracy and macro-F1 values are unchanged and
remain higher than those of RCProb in this six-dataset study.

\input{tables/t6_calibration}

The higher-capacity Dirichlet-L2 analysis is reported in Table~S8. It brings
RC+ log-loss close to raw RCProb for both base
ensembles, but can change hard predictions. Applying post-hoc calibration to
RCProb itself also leaves measurable headroom. The controlled experiment
therefore shows that RCProb improves native probability quality without
implying that subsequent calibration is unnecessary.

\subsection{Decision-list and aggregate inference}
\label{sec:res-inference}

Decision-list inference remains primary because one selected rule supplies
both the local explanation and the probability vector. The standardized
rulesets contain overlapping antecedents, however. Among held-out instances
covered by at least one non-default rule, an RF instance matches 4.28 RC+
rules and 5.18 RCProb rules on average. The corresponding GBM values are
3.16 and 3.38 (Table~S11).

Under aggregate inference, mean log-loss is 1.6053 for RC+ and 0.4506 for
RCProb in RF, and 1.4278 and 0.4750, respectively, in GBM (Table~S10). To
isolate the inference operation from ruleset selection, we also examined
cells where decision-list and aggregate HPO selected exactly the same stored
ruleset. In RF, aggregate inference reduces RC+ log-loss from 1.700 to
1.424, compared with 0.456 to 0.447 for RCProb. In GBM, the corresponding
changes are 1.475 to 1.231 and 0.457 to 0.446 (Table~S12). Macro-F1 changes
little in these identical-ruleset cells. The full aggregate results are
provided in Table~S10 and the rule-use analysis in Table~S11.

%% file: tables/t3_rf_main.tex
\begin{table*}[htbp]
\centering
\caption{Results for the random forest (RF) base ensemble. Each entry is the mean of the 23 dataset means with the standard deviation across those means; each dataset mean averages three repetitions and five outer folds. The source ensemble and the decision tree are shown as context. Bold values are the best among the four rule-extraction methods for that measure.}
\label{tab:rf-main}
\small
\resizebox{\textwidth}{!}{%
\begin{tabular}{lrrrrrr}
\toprule
\multicolumn{7}{l}{\textit{A. Probability quality, calibration, classification, and fidelity}} \\
Method & Log-loss $\downarrow$ & Brier $\downarrow$ & Conf.-ECE $\downarrow$ & cwECE $\downarrow$ & Macro-F1 $\uparrow$ & Fidelity $\uparrow$ \\
\midrule
Ensemble & $0.2971 \pm 0.26$ & $0.1681 \pm 0.14$ & $0.0682 \pm 0.04$ & $0.0656 \pm 0.04$ & $0.8513 \pm 0.13$ & -- \\
DT & $1.6842 \pm 1.85$ & $0.2255 \pm 0.19$ & $0.0746 \pm 0.07$ & $0.0605 \pm 0.06$ & $0.8284 \pm 0.14$ & -- \\
\addlinespace
RCProb & $\mathbf{0.4428} \pm 0.37$ & $\mathbf{0.2096} \pm 0.17$ & $0.0623 \pm 0.04$ & $0.0545 \pm 0.04$ & $0.8278 \pm 0.15$ & $\mathbf{0.9161} \pm 0.09$ \\
RC+ & $1.9003 \pm 1.89$ & $0.2278 \pm 0.20$ & $0.0895 \pm 0.08$ & $0.0707 \pm 0.06$ & $\mathbf{0.8339} \pm 0.14$ & $0.9088 \pm 0.09$ \\
inTrees & $0.7207 \pm 0.75$ & $0.2517 \pm 0.18$ & $0.0679 \pm 0.05$ & $0.0616 \pm 0.05$ & $0.7784 \pm 0.17$ & $0.8820 \pm 0.11$ \\
DFT & $0.6561 \pm 0.41$ & $0.3511 \pm 0.20$ & $\mathbf{0.0398} \pm 0.02$ & $\mathbf{0.0391} \pm 0.02$ & $0.5447 \pm 0.22$ & $0.7615 \pm 0.16$ \\
\midrule
\multicolumn{7}{l}{\textit{B. Ruleset structure}} \\
Method & Rules $\downarrow$ & Total ant. $\downarrow$ & Distinct ant. $\downarrow$ & Cond./rule $\downarrow$ & \textsc{uniq} $\uparrow$ & Fit time (s) $\downarrow$ \\
\midrule
DT & $27.83 \pm 34.45$ & $214.66 \pm 385.31$ & $51.74 \pm 64.09$ & $4.53 \pm 2.18$ & $0.472 \pm 0.16$ & $0.01 \pm 0.01$ \\
\addlinespace
RCProb & $20.00 \pm 13.78$ & $111.91 \pm 124.94$ & $45.09 \pm 36.34$ & $4.36 \pm 2.20$ & $0.569 \pm 0.19$ & $13.31 \pm 22.54$ \\
RC+ & $35.82 \pm 27.29$ & $149.03 \pm 152.63$ & $74.26 \pm 61.27$ & $3.36 \pm 1.15$ & $0.634 \pm 0.16$ & $13.03 \pm 24.39$ \\
inTrees & $14.89 \pm 12.43$ & $42.71 \pm 40.52$ & $35.99 \pm 31.29$ & $\mathbf{2.64} \pm 0.54$ & $0.899 \pm 0.09$ & $22.02 \pm 43.03$ \\
DFT & $\mathbf{3.20} \pm 1.03$ & $\mathbf{18.24} \pm 17.34$ & $\mathbf{17.51} \pm 16.41$ & $4.86 \pm 3.75$ & $\mathbf{0.906} \pm 0.09$ & $\mathbf{6.20} \pm 16.73$ \\
\bottomrule
\end{tabular}%
}
\end{table*}

%% file: tables/t3_gbm_main.tex
\begin{table*}[htbp]
\centering
\caption{Results for the gradient boosting machine (GBM) base ensemble. Each entry is the mean of the 23 dataset means with the standard deviation across those means; each dataset mean averages three repetitions and five outer folds. The source ensemble and the decision tree are shown as context. Bold values are the best among the four rule-extraction methods for that measure.}
\label{tab:gbm-main}
\small
\resizebox{\textwidth}{!}{%
\begin{tabular}{lrrrrrr}
\toprule
\multicolumn{7}{l}{\textit{A. Probability quality, calibration, classification, and fidelity}} \\
Method & Log-loss $\downarrow$ & Brier $\downarrow$ & Conf.-ECE $\downarrow$ & cwECE $\downarrow$ & Macro-F1 $\uparrow$ & Fidelity $\uparrow$ \\
\midrule
Ensemble & $0.4238 \pm 0.41$ & $0.1779 \pm 0.16$ & $0.0639 \pm 0.06$ & $0.0552 \pm 0.05$ & $0.8591 \pm 0.12$ & -- \\
DT & $1.6842 \pm 1.85$ & $0.2255 \pm 0.19$ & $0.0746 \pm 0.07$ & $0.0605 \pm 0.06$ & $0.8284 \pm 0.14$ & -- \\
\addlinespace
RCProb & $\mathbf{0.4935} \pm 0.44$ & $\mathbf{0.2100} \pm 0.17$ & $0.0571 \pm 0.04$ & $0.0519 \pm 0.04$ & $0.8287 \pm 0.14$ & $0.9104 \pm 0.09$ \\
RC+ & $1.6947 \pm 1.72$ & $0.2232 \pm 0.19$ & $0.0812 \pm 0.08$ & $0.0684 \pm 0.07$ & $\mathbf{0.8378} \pm 0.14$ & $\mathbf{0.9200} \pm 0.08$ \\
inTrees & $0.7751 \pm 0.75$ & $0.2599 \pm 0.19$ & $0.0731 \pm 0.05$ & $0.0681 \pm 0.05$ & $0.7671 \pm 0.18$ & $0.8648 \pm 0.12$ \\
DFT & $0.6442 \pm 0.41$ & $0.3616 \pm 0.20$ & $\mathbf{0.0424} \pm 0.02$ & $\mathbf{0.0392} \pm 0.02$ & $0.5367 \pm 0.21$ & $0.7294 \pm 0.17$ \\
\midrule
\multicolumn{7}{l}{\textit{B. Ruleset structure}} \\
Method & Rules $\downarrow$ & Total ant. $\downarrow$ & Distinct ant. $\downarrow$ & Cond./rule $\downarrow$ & \textsc{uniq} $\uparrow$ & Fit time (s) $\downarrow$ \\
\midrule
DT & $27.83 \pm 34.45$ & $214.66 \pm 385.31$ & $51.74 \pm 64.09$ & $4.53 \pm 2.18$ & $0.472 \pm 0.16$ & $0.01 \pm 0.01$ \\
\addlinespace
RCProb & $14.53 \pm 9.08$ & $72.07 \pm 88.31$ & $33.44 \pm 27.28$ & $3.80 \pm 2.09$ & $0.651 \pm 0.20$ & $7.92 \pm 11.51$ \\
RC+ & $25.70 \pm 17.97$ & $95.81 \pm 91.02$ & $55.48 \pm 44.08$ & $3.06 \pm 1.06$ & $0.697 \pm 0.15$ & $6.76 \pm 9.41$ \\
inTrees & $13.73 \pm 10.28$ & $35.25 \pm 29.83$ & $29.67 \pm 22.81$ & $\mathbf{2.40} \pm 0.52$ & $\mathbf{0.897} \pm 0.08$ & $14.45 \pm 26.79$ \\
DFT & $\mathbf{3.25} \pm 1.05$ & $\mathbf{16.43} \pm 13.40$ & $\mathbf{15.65} \pm 12.93$ & $4.48 \pm 3.76$ & $0.890 \pm 0.11$ & $\mathbf{3.94} \pm 8.23$ \\
\bottomrule
\end{tabular}%
}
\end{table*}

%% file: tables/s9_primary_tests.tex
\begin{table*}[htbp]
\centering
\caption{Primary RCProb--RC+ statistical comparison reported separately for RF and GBM. Each test uses 23 paired dataset means. Within each base ensemble, Holm's procedure adjusts the same 12 outcomes. $\Delta$ is the median paired relative change from RC+ to RCProb.}
\label{tab:primary-tests}
\scriptsize
\resizebox{\textwidth}{!}{%
\begin{tabular}{lrrrrrr}
\toprule
 & \multicolumn{3}{c}{RF} & \multicolumn{3}{c}{GBM} \\
\cmidrule(lr){2-4}\cmidrule(lr){5-7}
Measure & $\Delta$ (\%) & $p$ & $p_{\mathrm{Holm}}$ & $\Delta$ (\%) & $p$ & $p_{\mathrm{Holm}}$ \\
\midrule
Log-loss $\downarrow$ & -71.9 & $<0.0001$ & $<0.0001$ & -62.5 & $<0.0001$ & $<0.0001$ \\
Brier score $\downarrow$ & -3.7 & 0.1114 & 0.3920 & +1.0 & 0.6010 & 0.6010 \\
Confidence-ECE $\downarrow$ & -25.4 & 0.0060 & 0.0362 & -20.2 & 0.0075 & 0.0596 \\
Classwise-ECE $\downarrow$ & -15.5 & 0.0149 & 0.0743 & -18.2 & 0.0484 & 0.1453 \\
Macro-F1 $\uparrow$ & -0.5 & 0.3931 & 0.7530 & -0.7 & 0.0980 & 0.1960 \\
Fidelity $\uparrow$ & +0.3 & 0.3765 & 0.7530 & -0.6 & 0.0233 & 0.1369 \\
Rules $\downarrow$ & -38.7 & $<0.0001$ & $<0.0001$ & -38.5 & $<0.0001$ & 0.0006 \\
Total antecedents $\downarrow$ & -20.1 & 0.0027 & 0.0191 & -27.1 & 0.0004 & 0.0032 \\
Distinct antecedents $\downarrow$ & -34.8 & $<0.0001$ & $<0.0001$ & -32.7 & $<0.0001$ & $<0.0001$ \\
Conditions/rule $\downarrow$ & +17.3 & 0.0013 & 0.0102 & +15.7 & 0.0196 & 0.1369 \\
\textsc{uniq} $\uparrow$ & -11.6 & 0.0006 & 0.0057 & -6.6 & 0.0228 & 0.1369 \\
Fit time (s) $\downarrow$ & +25.3 & 0.0980 & 0.3920 & +39.8 & 0.0327 & 0.1369 \\
\bottomrule
\end{tabular}%
}
\end{table*}

%% file: tables/t7_case_cryotherapy.tex
\begin{table*}[htbp]
\centering
\caption{Illustrative \textit{cryotherapy}--GBM example selected because RC+ and RCProb retain the same first rule, which isolates the probability-estimation difference. The example illustrates a potential use of RCProb and is not selected to represent average performance; overall GBM benchmark results are reported in Table~\ref{tab:gbm-main}.}
\label{tab:case-cryotherapy}
\small
\resizebox{\textwidth}{!}{%
\begin{tabular}{p{0.24\textwidth}p{0.48\textwidth}rr}
\toprule
 & & RC+ & RCProb \\
\midrule
Identical rule & $(\mathrm{Time}\leq 8.125)\wedge(\mathrm{age}\leq 50)\rightarrow 1$ & & \\
Stated class-1 probability & & 1.000 & 0.956 \\
Held-out class-1 frequency & & 0.857 (6/7) & 0.857 (6/7) \\
\addlinespace\midrule
Fold log-loss & & 4.649 & 0.481 \\
Fold Brier score & & 0.366 & 0.287 \\
Fold confidence-ECE & & 0.197 & 0.078 \\
Fold macro-F1 & & 0.778 & 0.833 \\
Fidelity & & 1.000 & 0.944 \\
Rules & & 7 & 5 \\
Total antecedents & & 12 & 6 \\
\bottomrule
\end{tabular}%
}
\end{table*}

%% file: tables/t8_case_rules_rcplus.tex
\begin{table*}[htbp]
\centering
\caption{Final RC+ ruleset for the illustrative \textit{cryotherapy}--GBM fold. Coverage and training precision are given in percent, samples gives the class counts $[N_{r,0},N_{r,1}]$, and held-out counts the records for which the rule fired first.}
\label{tab:case-rules-rcplus}
\small
\setlength{\tabcolsep}{4pt}
\resizebox{\textwidth}{!}{%
\begin{tabular}{lrrrlp{0.40\textwidth}cll}
\toprule
 & $\mathrm{cov}$ & $\mathrm{prec}$ & $r.p_{r.y}$ & samples & $r.A$ & & $r.y$ & held-out \\
\midrule
$r_{1}$ & 45.83 & 100.00 & 1.000 & $[0,33]$ & $(\mathrm{Time} \leq 8.125) \wedge (\mathrm{age} \leq 50)$ & $\rightarrow$ & $[1]$ & 6/7 \\
$r_{2}$ & 30.56 & 100.00 & 1.000 & $[22,0]$ & $(\mathrm{Time} > 8.125) \wedge (\mathrm{age} > 16.5) \wedge (\mathrm{Area} > 25)$ & $\rightarrow$ & $[0]$ & 6/8 \\
$r_{3}$ & 11.11 & 100.00 & 1.000 & $[8,0]$ & $(\mathrm{Time} > 8.125) \wedge (\mathrm{Area} \leq 15)$ & $\rightarrow$ & $[0]$ & 1/1 \\
$r_{4}$ & 6.94 & 100.00 & 1.000 & $[0,5]$ & $(\mathrm{Area} > 15) \wedge (\mathrm{Type} \leq 2.5) \wedge (\mathrm{Area} \leq 25)$ & $\rightarrow$ & $[1]$ & -- \\
$r_{5}$ & 8.33 & 100.00 & 1.000 & $[6,0]$ & $(\mathrm{age} > 50)$ & $\rightarrow$ & $[0]$ & -- \\
$r_{6}$ & 16.67 & 100.00 & 1.000 & $[0,12]$ & $(\mathrm{age} \leq 16.5)$ & $\rightarrow$ & $[1]$ & 1/1 \\
$r_d$ & 0.00 & -- & 0.542 & $[0,0]$ & $(\,)$ & $\rightarrow$ & $[1]$ & 0/1 \\
\bottomrule
\end{tabular}%
}
\end{table*}

%% file: tables/t9_case_rules_rcprob.tex
\begin{table*}[htbp]
\centering
\caption{Final RCProb ruleset for the illustrative \textit{cryotherapy}--GBM fold. Coverage and training precision are given in percent, samples gives the class counts $[N_{r,0},N_{r,1}]$, and held-out counts the records for which the rule fired first.}
\label{tab:case-rules-rcprob}
\small
\setlength{\tabcolsep}{4pt}
\resizebox{\textwidth}{!}{%
\begin{tabular}{lrrrlp{0.40\textwidth}cll}
\toprule
 & $\mathrm{cov}$ & $\mathrm{prec}$ & $r.p_{r.y}$ & samples & $r.A$ & & $r.y$ & held-out \\
\midrule
$r_{1}$ & 45.83 & 100.00 & 0.956 & $[0,33]$ & $(\mathrm{Time} \leq 8.125) \wedge (\mathrm{age} \leq 50)$ & $\rightarrow$ & $[1]$ & 6/7 \\
$r_{2}$ & 45.83 & 90.91 & 0.875 & $[30,3]$ & $(\mathrm{Time} > 8.125) \wedge (\mathrm{age} > 16.5)$ & $\rightarrow$ & $[0]$ & 8/10 \\
$r_{3}$ & 8.33 & 100.00 & 0.939 & $[6,0]$ & $(\mathrm{age} > 50)$ & $\rightarrow$ & $[0]$ & -- \\
$r_{4}$ & 16.67 & 100.00 & 0.958 & $[0,12]$ & $(\mathrm{age} \leq 16.5)$ & $\rightarrow$ & $[1]$ & 1/1 \\
$r_d$ & 0.00 & -- & 0.542 & $[0,0]$ & $(\,)$ & $\rightarrow$ & $[1]$ & -- \\
\bottomrule
\end{tabular}%
}
\end{table*}

%% file: tables/t6_calibration.tex
\begin{table*}[htbp]
\centering
\caption{Controlled post-hoc calibration experiment reported separately for RF and GBM. Each entry is the descriptive mean over six datasets with the standard deviation across those dataset means, each averaging five repeated 60/20/20 development/calibration/test splits. Temperature scaling is fitted only on the dedicated calibration partition and preserves the predicted class.}
\label{tab:calibration}
\small
\resizebox{\textwidth}{!}{%
\begin{tabular}{lrrrrr}
\toprule
\multicolumn{6}{l}{\textit{RF}} \\
Measure & RC+ & RC+ + TempS & DT & DT + TempS & RCProb \\
\midrule
Log-loss $\downarrow$ & $1.1392 \pm 1.98$ & $0.2904 \pm 0.30$ & $1.5996 \pm 3.10$ & $0.3194 \pm 0.35$ & $0.2482 \pm 0.25$ \\
Brier score $\downarrow$ & $0.1293 \pm 0.19$ & $0.1267 \pm 0.16$ & $0.1573 \pm 0.23$ & $0.1456 \pm 0.19$ & $0.1183 \pm 0.15$ \\
Confidence-ECE $\downarrow$ & $0.0571 \pm 0.09$ & $0.0452 \pm 0.05$ & $0.0638 \pm 0.11$ & $0.0388 \pm 0.04$ & $0.0335 \pm 0.03$ \\
Classwise-ECE $\downarrow$ & $0.0373 \pm 0.04$ & $0.0522 \pm 0.05$ & $0.0394 \pm 0.05$ & $0.0506 \pm 0.04$ & $0.0276 \pm 0.02$ \\
Accuracy $\uparrow$ & $0.9266 \pm 0.11$ & $0.9266 \pm 0.11$ & $0.9148 \pm 0.12$ & $0.9148 \pm 0.12$ & $0.9197 \pm 0.11$ \\
Macro-F1 $\uparrow$ & $0.8784 \pm 0.10$ & $0.8784 \pm 0.10$ & $0.8650 \pm 0.11$ & $0.8650 \pm 0.11$ & $0.8653 \pm 0.11$ \\
\midrule
\multicolumn{6}{l}{\textit{GBM}} \\
Measure & RC+ & RC+ + TempS & DT & DT + TempS & RCProb \\
\midrule
Log-loss $\downarrow$ & $0.9674 \pm 1.65$ & $0.2691 \pm 0.28$ & $1.5996 \pm 3.10$ & $0.3194 \pm 0.35$ & $0.2354 \pm 0.26$ \\
Brier score $\downarrow$ & $0.1215 \pm 0.17$ & $0.1183 \pm 0.15$ & $0.1573 \pm 0.23$ & $0.1456 \pm 0.19$ & $0.1175 \pm 0.15$ \\
Confidence-ECE $\downarrow$ & $0.0474 \pm 0.07$ & $0.0337 \pm 0.03$ & $0.0638 \pm 0.11$ & $0.0388 \pm 0.04$ & $0.0272 \pm 0.03$ \\
Classwise-ECE $\downarrow$ & $0.0315 \pm 0.04$ & $0.0412 \pm 0.04$ & $0.0394 \pm 0.05$ & $0.0506 \pm 0.04$ & $0.0221 \pm 0.02$ \\
Accuracy $\uparrow$ & $0.9288 \pm 0.10$ & $0.9288 \pm 0.10$ & $0.9148 \pm 0.12$ & $0.9148 \pm 0.12$ & $0.9189 \pm 0.11$ \\
Macro-F1 $\uparrow$ & $0.8838 \pm 0.09$ & $0.8838 \pm 0.09$ & $0.8650 \pm 0.11$ & $0.8650 \pm 0.11$ & $0.8660 \pm 0.11$ \\
\bottomrule
\end{tabular}%
}
\end{table*}

%% file: sections/07_discussion.tex
\section{Discussion}
\label{sec:discussion}

\subsection{Why the probability estimates change}
\label{sec:disc-mechanism}

A plausible explanation starts with how rule precision is used inside a
greedy search. RC+ evaluates many candidate conjunctions and retains
candidates partly because their observed training precision is high. When a
noisy estimate is also used to select among alternatives, the selected
estimates can be optimistic even when the underlying estimator is unbiased.
This is the same general selection effect described by the winner's curse
literature \citep{andrewsInferenceWinners2024}.
\citet{jensenMultipleComparisonsInduction2000} attribute overfitting and
oversearching in induction algorithms to this mechanism and show that the
maximum sample score is a positively biased estimate of the population score
of the selected item. The present experiment does not isolate post-selection
bias as a causal mechanism, so this interpretation remains an explanation of
the observed pattern, not a separately tested result.

The held-out rule diagnostics are consistent with that explanation for both
RF and GBM. More than 70\% of RC+ non-default rules state a probability of
exactly 1.0 in each base-ensemble analysis, whereas RCProb produces none.
The largest descriptive subgroup gaps occur among rules assigned only a few
held-out observations. RCProb changes both the statistic used during the
search stages and the probability attached after exact support is
recomputed.

Naive Bayes probabilities can themselves be overconfident when the
conditional-independence approximation is violated
\citep{silva2023classifier,kullBetaCalibrationWellfounded2017}. RCProb
therefore uses the Naive Bayes posterior as an inexpensive composition of
atomic evidence, and leaves the regularization of the final rule probability
to the steps that follow. Additive smoothing prevents zero or one atomic
likelihoods, and the final rule probability mixes this estimate with an
ensemble-informed posterior. The support-adaptive weight gives the
m-estimate its largest influence for sparse rules and increases the Naive
Bayes weight as support grows. The targeted sensitivity analysis in
Table~S13 is consistent with this choice. Increasing $n_0$ beyond 50, which
retains more prior weight at a given support, does not improve F1 or
rule count and progressively increases log-loss in both base ensembles. The
probability result therefore belongs to the complete hybrid procedure, not
to Naive Bayes alone.

\subsection{Calibration error and probability quality are different}
\label{sec:disc-metrics}

The results also show why calibration-specific measures and proper scoring
rules should be reported together. The calibration survey of
\citet{silva2023classifier} distinguishes confidence calibration, classwise
calibration, and stronger multiclass notions, and notes that ECE can be
small for uninformative probability forecasts. Proper scoring rules such as
log-loss and Brier score evaluate the complete probabilistic prediction and
cannot be replaced by ECE alone.

This distinction appears in both base-ensemble comparisons. DFT has the
lowest numerical confidence-ECE and classwise-ECE among the interpretable
models, but its macro-F1, fidelity, Brier score, and log-loss are much
worse. Figure~\ref{fig:reliability-methods} shows how it obtains that
calibration. DFT states a mean probability of 0.7229 at a held-out accuracy
of 0.7166, so its stated probabilities are close to correct because they are
low, and only 11\% of a dataset's observations are stated above 0.99. RCProb
states 0.9096 at an accuracy of 0.8605, with 35\% above 0.99.
The five-method Friedman tests do not establish confidence-ECE
differences for either RF or GBM, which further argues against selecting a
method from ECE alone. The source ensembles retain better proper-scoring and
classification performance than the simplified models. RCProb therefore
recovers part of the probabilistic information lost in simplification, and
its main effect is to reduce the severe degradation observed after RC+
extraction.

\subsection{Post-hoc calibration and probabilistic rule extraction}
\label{sec:disc-posthoc}

The controlled experiment shows that a conventional post-hoc calibrator is a
strong alternative when the main objective is model-level probability
quality. Temperature scaling repairs much of RC+'s log-loss for both RF and
GBM without changing any hard classification decision. The more flexible
Dirichlet calibrator reduces it further. Much of the RC+ probability problem
can therefore be repaired after extraction, given held-out data on which to
fit the calibration map.

RCProb addresses a different design point. Its native probabilities are
competitive with calibrated RC+ without fitting a separate calibration map
or reserving a calibration partition for deployment of the ruleset. The
probability estimate is also available while the rules are being searched,
so it can affect which candidates survive. Post-hoc calibration operates
only after the ruleset has been selected. The controlled experiment also
shows that RCProb can benefit from subsequent calibration. Where calibrated
probabilities are a deployment requirement, the two remain complementary.

\subsection{Global rulesets, local explanations, and overlapping rules}
\label{sec:disc-inference}

The held-out use analysis shows substantial antecedent overlap in both RF and
GBM, so several non-default rules can match one instance while only the
first is selected, and rules that never fire first remain part of the global
representation and can contribute under aggregate inference.

Aggregation reduces RC+ log-loss in both base ensembles, including when the
stored ruleset is held fixed. The corresponding change for RCProb is much
smaller. Averaging several overlapping RC+ probability vectors therefore
softens part of the extreme probability behavior. The cost is a less direct
local explanation because the final probability is no longer attributable to
one rule. RCProb is most directly useful when the decision-list rule is
intended to provide both the local explanation and its probability
statement.

%% file: sections/08_conclusion.tex
\section{Conclusion}
\label{sec:conclusion}

In this paper, we presented RCProb, a probabilistic rule extraction method
for classification tree ensembles. The method retains the RC+ combination and
simplification framework, and replaces the training precision that RC+ uses
to score and report a rule with smoothed atomic evidence during the search
stages and a support-adaptive hybrid probability on the final rules.

Several experiments were performed to evaluate and validate the results,
using two base ensemble algorithms, three benchmark ensemble simplification
methods and a single decision tree as an interpretable reference. Relative to
RC+, the median paired log-loss reduction is 71.9\% for RF and 62.5\% for
GBM, and the number of rules falls by 38.7\% and 38.5\%, with all four
differences remaining significant after Holm correction. Confidence-ECE is
also lower in both base ensembles, and the reduction is significant for RF.
The primary tests do not detect a macro-F1 difference in either case,
although the retained rules are longer on average. Among the comparison
methods, inTrees and DFT returned more compact rulesets, while RCProb
performed substantially better on both probabilistic and predictive
measures. At the rule level, RCProb removes the near-certain probability
values that dominate RC+ rulesets, which matters under decision-list
inference because the selected rule supplies both the local explanation and
its probability statement.

This study has some limitations. The experiments cover classification only,
and the sizes of the datasets were limited to those used in the experiments.
The controlled calibration comparison uses six datasets and is therefore
descriptive. The proposed method can be extended in different ways. Adding
the regression task is one direction, and it is not trivial, because the
combination process needs a mechanism for merging predicted values. It would
also be worth comparing alternative rules for combining the local priors of
merged rules, and investigating probability estimators for overlapping
rulesets where several rules jointly determine a prediction.

%% file: tables/s1_per_dataset_rf.tex
\begin{table*}[htbp]
\centering
\caption{Dataset-level means for the primary RC+--RCProb comparison with RF. Each entry averages three repetitions and five outer folds.}
\label{tab:supp-perdataset-rf}
\scriptsize
\resizebox{\textwidth}{!}{%
\begin{tabular}{lrrrrrrrrrr}
\toprule
 & \multicolumn{2}{c}{Macro-F1} & \multicolumn{2}{c}{Log-loss} & \multicolumn{2}{c}{Confidence-ECE} & \multicolumn{2}{c}{Brier} & \multicolumn{2}{c}{Rules} \\
\cmidrule(lr){2-3}\cmidrule(lr){4-5}\cmidrule(lr){6-7}\cmidrule(lr){8-9}\cmidrule(lr){10-11}
Dataset & RC+ & RCProb & RC+ & RCProb & RC+ & RCProb & RC+ & RCProb & RC+ & RCProb \\
\midrule
australian & 0.844 & 0.849 & 1.532 & 0.426 & 0.109 & 0.078 & 0.256 & 0.241 & 35.5 & 14.9 \\
bands & 0.638 & 0.654 & 6.384 & 0.652 & 0.267 & 0.130 & 0.580 & 0.444 & 38.6 & 21.0 \\
banknote & 0.989 & 0.957 & 0.240 & 0.149 & 0.010 & 0.028 & 0.021 & 0.074 & 25.7 & 11.9 \\
biodeg & 0.826 & 0.804 & 2.500 & 0.541 & 0.128 & 0.114 & 0.277 & 0.300 & 65.3 & 25.6 \\
cardiotoc & 0.914 & 0.883 & 0.936 & 0.263 & 0.049 & 0.043 & 0.106 & 0.129 & 54.0 & 34.0 \\
credit & 0.990 & 0.985 & 0.087 & 0.053 & 0.003 & 0.005 & 0.006 & 0.009 & 3.4 & 3.9 \\
cryotherapy & 0.810 & 0.882 & 3.881 & 0.487 & 0.141 & 0.088 & 0.309 & 0.200 & 8.7 & 7.8 \\
divorce & 0.965 & 0.976 & 1.002 & 0.189 & 0.035 & 0.024 & 0.071 & 0.045 & 5.3 & 4.5 \\
ecoli & 0.744 & 0.761 & 2.605 & 0.800 & 0.122 & 0.080 & 0.310 & 0.258 & 25.0 & 17.5 \\
glass & 0.600 & 0.514 & 6.454 & 1.454 & 0.260 & 0.153 & 0.563 & 0.543 & 25.2 & 12.1 \\
heart & 0.790 & 0.810 & 2.872 & 0.524 & 0.149 & 0.111 & 0.357 & 0.307 & 17.8 & 14.1 \\
htru2 & 0.934 & 0.930 & 0.116 & 0.147 & 0.012 & 0.018 & 0.038 & 0.041 & 48.6 & 29.8 \\
ionosphere & 0.897 & 0.920 & 1.817 & 0.295 & 0.083 & 0.052 & 0.177 & 0.136 & 11.9 & 10.2 \\
maintenance & 0.836 & 0.756 & 0.143 & 0.099 & 0.014 & 0.012 & 0.035 & 0.046 & 59.4 & 41.3 \\
mammographic & 0.826 & 0.826 & 0.474 & 0.482 & 0.068 & 0.090 & 0.264 & 0.279 & 11.5 & 10.5 \\
new-thyroid & 0.910 & 0.930 & 1.392 & 0.303 & 0.063 & 0.043 & 0.126 & 0.094 & 9.4 & 8.6 \\
occupancy & 0.989 & 0.983 & 0.077 & 0.058 & 0.005 & 0.009 & 0.013 & 0.022 & 23.8 & 12.3 \\
pageblocks0 & 0.927 & 0.909 & 0.225 & 0.113 & 0.018 & 0.015 & 0.046 & 0.054 & 52.3 & 30.1 \\
pima & 0.692 & 0.719 & 2.670 & 0.562 & 0.145 & 0.101 & 0.415 & 0.365 & 47.6 & 23.5 \\
spambase & 0.943 & 0.925 & 0.794 & 0.315 & 0.040 & 0.054 & 0.099 & 0.128 & 114.9 & 62.4 \\
vehicle & 0.711 & 0.681 & 4.203 & 0.788 & 0.197 & 0.097 & 0.463 & 0.424 & 55.9 & 19.8 \\
wisconsin & 0.953 & 0.956 & 0.567 & 0.202 & 0.032 & 0.033 & 0.075 & 0.076 & 10.9 & 10.8 \\
yeast & 0.451 & 0.429 & 2.735 & 1.282 & 0.110 & 0.056 & 0.631 & 0.604 & 73.1 & 33.6 \\
\bottomrule
\end{tabular}%
}
\end{table*}

%% file: tables/s1_per_dataset_gbm.tex
\begin{table*}[htbp]
\centering
\caption{Dataset-level means for the primary RC+--RCProb comparison with GBM. Each entry averages three repetitions and five outer folds.}
\label{tab:supp-perdataset-gbm}
\scriptsize
\resizebox{\textwidth}{!}{%
\begin{tabular}{lrrrrrrrrrr}
\toprule
 & \multicolumn{2}{c}{Macro-F1} & \multicolumn{2}{c}{Log-loss} & \multicolumn{2}{c}{Confidence-ECE} & \multicolumn{2}{c}{Brier} & \multicolumn{2}{c}{Rules} \\
\cmidrule(lr){2-3}\cmidrule(lr){4-5}\cmidrule(lr){6-7}\cmidrule(lr){8-9}\cmidrule(lr){10-11}
Dataset & RC+ & RCProb & RC+ & RCProb & RC+ & RCProb & RC+ & RCProb & RC+ & RCProb \\
\midrule
australian & 0.845 & 0.856 & 1.060 & 0.394 & 0.076 & 0.060 & 0.239 & 0.225 & 19.5 & 10.2 \\
bands & 0.629 & 0.644 & 6.447 & 0.678 & 0.297 & 0.137 & 0.615 & 0.458 & 32.7 & 19.9 \\
banknote & 0.986 & 0.963 & 0.320 & 0.140 & 0.012 & 0.030 & 0.027 & 0.066 & 19.3 & 10.8 \\
biodeg & 0.825 & 0.818 & 2.496 & 0.481 & 0.112 & 0.089 & 0.269 & 0.272 & 47.4 & 22.1 \\
cardiotoc & 0.916 & 0.891 & 0.761 & 0.237 & 0.040 & 0.029 & 0.100 & 0.122 & 39.2 & 22.1 \\
credit & 0.994 & 0.991 & 0.055 & 0.017 & 0.002 & 0.006 & 0.004 & 0.006 & 3.0 & 3.0 \\
cryotherapy & 0.861 & 0.857 & 3.653 & 1.370 & 0.139 & 0.098 & 0.270 & 0.249 & 7.7 & 6.7 \\
divorce & 0.972 & 0.976 & 0.758 & 0.118 & 0.027 & 0.026 & 0.055 & 0.046 & 4.5 & 4.5 \\
ecoli & 0.761 & 0.755 & 2.062 & 0.929 & 0.096 & 0.077 & 0.279 & 0.272 & 16.5 & 12.7 \\
glass & 0.641 & 0.583 & 4.788 & 1.657 & 0.182 & 0.142 & 0.510 & 0.512 & 19.9 & 12.3 \\
heart & 0.757 & 0.804 & 3.732 & 0.529 & 0.182 & 0.101 & 0.418 & 0.311 & 18.9 & 11.6 \\
htru2 & 0.932 & 0.933 & 0.094 & 0.108 & 0.008 & 0.015 & 0.039 & 0.041 & 20.4 & 13.5 \\
ionosphere & 0.884 & 0.910 & 2.496 & 0.297 & 0.101 & 0.057 & 0.208 & 0.140 & 13.1 & 10.4 \\
maintenance & 0.845 & 0.751 & 0.144 & 0.101 & 0.010 & 0.008 & 0.031 & 0.047 & 53.5 & 18.4 \\
mammographic & 0.828 & 0.826 & 0.439 & 0.464 & 0.058 & 0.080 & 0.266 & 0.275 & 8.3 & 6.6 \\
new-thyroid & 0.917 & 0.910 & 1.201 & 0.529 & 0.050 & 0.046 & 0.108 & 0.126 & 8.5 & 7.9 \\
occupancy & 0.989 & 0.983 & 0.086 & 0.060 & 0.006 & 0.009 & 0.014 & 0.022 & 21.3 & 8.1 \\
pageblocks0 & 0.917 & 0.911 & 0.235 & 0.110 & 0.017 & 0.011 & 0.051 & 0.054 & 40.5 & 24.3 \\
pima & 0.706 & 0.710 & 1.475 & 0.570 & 0.122 & 0.098 & 0.396 & 0.370 & 23.5 & 17.3 \\
spambase & 0.937 & 0.931 & 0.585 & 0.273 & 0.039 & 0.045 & 0.105 & 0.119 & 72.1 & 46.6 \\
vehicle & 0.713 & 0.679 & 3.494 & 0.776 & 0.180 & 0.068 & 0.454 & 0.421 & 52.1 & 16.5 \\
wisconsin & 0.954 & 0.956 & 0.733 & 0.263 & 0.034 & 0.033 & 0.075 & 0.078 & 13.9 & 11.9 \\
yeast & 0.457 & 0.425 & 1.864 & 1.252 & 0.074 & 0.048 & 0.601 & 0.595 & 35.2 & 16.9 \\
\bottomrule
\end{tabular}%
}
\end{table*}

%% file: tables/s10_secondary_rf.tex
\begin{table*}[htbp]
\centering
\caption{Secondary five-method benchmark for RF. The Friedman test compares RCProb, RC+, DT, inTrees, and DFT across 23 datasets. Where the omnibus test is significant, cells report the direction of RCProb relative to each comparator and the Holm-adjusted pairwise $p$-value across the four RCProb comparisons. B = RCProb better; W = RCProb worse; n.s. = not significant. A dash indicates that the omnibus test was not significant and post-hoc results are not interpreted.}
\label{tab:supp-secondary-rf}
\scriptsize
\resizebox{\textwidth}{!}{%
\begin{tabular}{lrrrrr}
\toprule
Measure & Friedman $p$ & RC+ & DT & inTrees & DFT \\
\midrule
Log-loss $\downarrow$ & $<0.0001$ & B ($<0.0001$) & B ($<0.0001$) & B ($<0.0001$) & B ($<0.0001$) \\
Brier score $\downarrow$ & $<0.0001$ & n.s. (0.2229) & n.s. (0.2229) & B ($<0.0001$) & B ($<0.0001$) \\
Confidence-ECE $\downarrow$ & 0.0866 & -- & -- & -- & -- \\
Classwise-ECE $\downarrow$ & 0.1240 & -- & -- & -- & -- \\
Macro-F1 $\uparrow$ & $<0.0001$ & n.s. (0.7861) & n.s. (0.7861) & B (0.0003) & B ($<0.0001$) \\
Fidelity $\uparrow$ & $<0.0001$ & n.s. (0.7530) & n.s. (0.7530) & B (0.0022) & B ($<0.0001$) \\
Rules $\downarrow$ & $<0.0001$ & B ($<0.0001$) & n.s. (0.5202) & n.s. (0.2009) & W ($<0.0001$) \\
Total antecedents $\downarrow$ & $<0.0001$ & B (0.0082) & n.s. (0.4274) & W (0.0121) & W ($<0.0001$) \\
Distinct antecedents $\downarrow$ & $<0.0001$ & B ($<0.0001$) & n.s. (0.6588) & n.s. (0.6588) & W ($<0.0001$) \\
Conditions/rule $\downarrow$ & $<0.0001$ & W (0.0038) & n.s. (1.0000) & W (0.0002) & n.s. (1.0000) \\
\textsc{uniq} $\uparrow$ & $<0.0001$ & W (0.0013) & B (0.0075) & W ($<0.0001$) & W ($<0.0001$) \\
Fit time (s) $\downarrow$ & $<0.0001$ & n.s. (0.1960) & W ($<0.0001$) & n.s. (0.6010) & W (0.0025) \\
\bottomrule
\end{tabular}%
}
\end{table*}

%% file: tables/s10_secondary_gbm.tex
\begin{table*}[htbp]
\centering
\caption{Secondary five-method benchmark for GBM. The Friedman test compares RCProb, RC+, DT, inTrees, and DFT across 23 datasets. Where the omnibus test is significant, cells report the direction of RCProb relative to each comparator and the Holm-adjusted pairwise $p$-value across the four RCProb comparisons. B = RCProb better; W = RCProb worse; n.s. = not significant. A dash indicates that the omnibus test was not significant and post-hoc results are not interpreted.}
\label{tab:supp-secondary-gbm}
\scriptsize
\resizebox{\textwidth}{!}{%
\begin{tabular}{lrrrrr}
\toprule
Measure & Friedman $p$ & RC+ & DT & inTrees & DFT \\
\midrule
Log-loss $\downarrow$ & $<0.0001$ & B ($<0.0001$) & B ($<0.0001$) & B ($<0.0001$) & B (0.0007) \\
Brier score $\downarrow$ & $<0.0001$ & n.s. (0.6010) & n.s. (0.2229) & B ($<0.0001$) & B ($<0.0001$) \\
Confidence-ECE $\downarrow$ & 0.2139 & -- & -- & -- & -- \\
Classwise-ECE $\downarrow$ & 0.0722 & -- & -- & -- & -- \\
Macro-F1 $\uparrow$ & $<0.0001$ & n.s. (0.1960) & n.s. (0.9406) & B ($<0.0001$) & B ($<0.0001$) \\
Fidelity $\uparrow$ & $<0.0001$ & W (0.0466) & n.s. (0.8462) & B ($<0.0001$) & B ($<0.0001$) \\
Rules $\downarrow$ & $<0.0001$ & B (0.0002) & n.s. (0.0602) & n.s. (0.6869) & W (0.0001) \\
Total antecedents $\downarrow$ & $<0.0001$ & B (0.0011) & B (0.0061) & W (0.0179) & W ($<0.0001$) \\
Distinct antecedents $\downarrow$ & $<0.0001$ & B ($<0.0001$) & n.s. (0.3390) & n.s. (0.7998) & W ($<0.0001$) \\
Conditions/rule $\downarrow$ & $<0.0001$ & n.s. (0.0587) & n.s. (0.1960) & W (0.0012) & n.s. (0.4452) \\
\textsc{uniq} $\uparrow$ & $<0.0001$ & W (0.0228) & B ($<0.0001$) & W (0.0002) & W (0.0007) \\
Fit time (s) $\downarrow$ & $<0.0001$ & n.s. (0.0653) & W ($<0.0001$) & n.s. (0.7998) & W (0.0303) \\
\bottomrule
\end{tabular}%
}
\end{table*}

%% file: tables/s11_ranks_rf.tex
\begin{table}[htbp]
\centering
\caption{Average ranks across the 23 datasets for the five-method RF benchmark. Rank 1 is best for the direction indicated by the metric arrow.}
\label{tab:supp-ranks-rf}
\scriptsize
\begin{tabular}{lrrrrr}
\toprule
Measure & RCProb & RC+ & DT & inTrees & DFT \\
\midrule
Log-loss $\downarrow$ & 1.39 & 4.13 & 3.91 & 2.43 & 3.13 \\
Brier score $\downarrow$ & 2.09 & 2.26 & 2.26 & 3.61 & 4.78 \\
Confidence-ECE $\downarrow$ & 2.87 & 3.74 & 2.96 & 3.00 & 2.43 \\
Classwise-ECE $\downarrow$ & 2.78 & 3.61 & 2.96 & 3.22 & 2.43 \\
Macro-F1 $\uparrow$ & 2.09 & 2.00 & 2.17 & 3.74 & 5.00 \\
Fidelity $\uparrow$ & 2.22 & 2.30 & 2.43 & 3.39 & 4.65 \\
Rules $\downarrow$ & 3.30 & 4.43 & 3.17 & 2.87 & 1.22 \\
Total antecedents $\downarrow$ & 3.48 & 4.04 & 3.61 & 2.35 & 1.52 \\
Distinct antecedents $\downarrow$ & 3.30 & 4.35 & 2.78 & 2.87 & 1.70 \\
Conditions/rule $\downarrow$ & 3.48 & 2.78 & 3.70 & 1.61 & 3.43 \\
\textsc{uniq} $\uparrow$ & 3.96 & 3.22 & 4.52 & 1.61 & 1.70 \\
Fit time (s) $\downarrow$ & 4.22 & 3.61 & 1.00 & 4.00 & 2.17 \\
\bottomrule
\end{tabular}
\end{table}

%% file: tables/s11_ranks_gbm.tex
\begin{table}[htbp]
\centering
\caption{Average ranks across the 23 datasets for the five-method GBM benchmark. Rank 1 is best for the direction indicated by the metric arrow.}
\label{tab:supp-ranks-gbm}
\scriptsize
\begin{tabular}{lrrrrr}
\toprule
Measure & RCProb & RC+ & DT & inTrees & DFT \\
\midrule
Log-loss $\downarrow$ & 1.30 & 3.83 & 4.04 & 2.57 & 3.26 \\
Brier score $\downarrow$ & 2.09 & 2.17 & 2.17 & 3.70 & 4.87 \\
Confidence-ECE $\downarrow$ & 2.57 & 3.30 & 3.00 & 3.48 & 2.65 \\
Classwise-ECE $\downarrow$ & 2.52 & 3.09 & 2.87 & 3.78 & 2.74 \\
Macro-F1 $\uparrow$ & 2.30 & 1.74 & 2.26 & 3.70 & 5.00 \\
Fidelity $\uparrow$ & 2.43 & 1.52 & 2.30 & 3.78 & 4.96 \\
Rules $\downarrow$ & 3.04 & 4.30 & 3.43 & 3.09 & 1.13 \\
Total antecedents $\downarrow$ & 3.04 & 4.17 & 4.13 & 2.22 & 1.43 \\
Distinct antecedents $\downarrow$ & 2.87 & 4.30 & 3.35 & 2.96 & 1.52 \\
Conditions/rule $\downarrow$ & 3.22 & 2.65 & 4.09 & 1.59 & 3.46 \\
\textsc{uniq} $\uparrow$ & 3.52 & 3.00 & 4.83 & 1.74 & 1.91 \\
Fit time (s) $\downarrow$ & 4.09 & 3.48 & 1.00 & 4.13 & 2.30 \\
\bottomrule
\end{tabular}
\end{table}

%% file: tables/s2_calibration_full.tex
\begin{table*}[htbp]
\centering
\caption{Full descriptive controlled-calibration results reported separately for RF and GBM. Each entry is the mean over the six dataset means with the standard deviation across them. TempS preserves the predicted class; Dirichlet-L2 may alter it.}
\label{tab:supp-calibration-full}
\scriptsize
\resizebox{\textwidth}{!}{%
\begin{tabular}{lrrrrrrrrr}
\toprule
\multicolumn{10}{l}{\textit{RF}} \\
Measure & RC+ & RC+ + TempS & RC+ + Dir-L2 & DT & DT + TempS & DT + Dir-L2 & RCProb & RCProb + TempS & RCProb + Dir-L2 \\
\midrule
Log-loss $\downarrow$ & $1.1392 \pm 1.98$ & $0.2904 \pm 0.30$ & $0.2467 \pm 0.25$ & $1.5996 \pm 3.10$ & $0.3194 \pm 0.35$ & $0.2666 \pm 0.28$ & $0.2482 \pm 0.25$ & $0.2215 \pm 0.26$ & $0.2395 \pm 0.30$ \\
Brier score $\downarrow$ & $0.1293 \pm 0.19$ & $0.1267 \pm 0.16$ & $0.1204 \pm 0.14$ & $0.1573 \pm 0.23$ & $0.1456 \pm 0.19$ & $0.1366 \pm 0.16$ & $0.1183 \pm 0.15$ & $0.1158 \pm 0.14$ & $0.1171 \pm 0.15$ \\
Confidence-ECE $\downarrow$ & $0.0571 \pm 0.09$ & $0.0452 \pm 0.05$ & $0.0432 \pm 0.05$ & $0.0638 \pm 0.11$ & $0.0388 \pm 0.04$ & $0.0432 \pm 0.05$ & $0.0335 \pm 0.03$ & $0.0199 \pm 0.03$ & $0.0222 \pm 0.04$ \\
Classwise-ECE $\downarrow$ & $0.0373 \pm 0.04$ & $0.0522 \pm 0.05$ & $0.0408 \pm 0.04$ & $0.0394 \pm 0.05$ & $0.0506 \pm 0.04$ & $0.0403 \pm 0.05$ & $0.0276 \pm 0.02$ & $0.0165 \pm 0.02$ & $0.0173 \pm 0.02$ \\
Accuracy $\uparrow$ & $0.9266 \pm 0.11$ & $0.9266 \pm 0.11$ & $0.9189 \pm 0.11$ & $0.9148 \pm 0.12$ & $0.9148 \pm 0.12$ & $0.9083 \pm 0.12$ & $0.9197 \pm 0.11$ & $0.9197 \pm 0.11$ & $0.9176 \pm 0.12$ \\
Macro-F1 $\uparrow$ & $0.8784 \pm 0.10$ & $0.8784 \pm 0.10$ & $0.8464 \pm 0.13$ & $0.8650 \pm 0.11$ & $0.8650 \pm 0.11$ & $0.8397 \pm 0.12$ & $0.8653 \pm 0.11$ & $0.8653 \pm 0.11$ & $0.8439 \pm 0.14$ \\
\midrule
\multicolumn{10}{l}{\textit{GBM}} \\
Measure & RC+ & RC+ + TempS & RC+ + Dir-L2 & DT & DT + TempS & DT + Dir-L2 & RCProb & RCProb + TempS & RCProb + Dir-L2 \\
\midrule
Log-loss $\downarrow$ & $0.9674 \pm 1.65$ & $0.2691 \pm 0.28$ & $0.2283 \pm 0.24$ & $1.5996 \pm 3.10$ & $0.3194 \pm 0.35$ & $0.2666 \pm 0.28$ & $0.2354 \pm 0.26$ & $0.2219 \pm 0.26$ & $0.2260 \pm 0.27$ \\
Brier score $\downarrow$ & $0.1215 \pm 0.17$ & $0.1183 \pm 0.15$ & $0.1100 \pm 0.13$ & $0.1573 \pm 0.23$ & $0.1456 \pm 0.19$ & $0.1366 \pm 0.16$ & $0.1175 \pm 0.15$ & $0.1161 \pm 0.15$ & $0.1183 \pm 0.15$ \\
Confidence-ECE $\downarrow$ & $0.0474 \pm 0.07$ & $0.0337 \pm 0.03$ & $0.0315 \pm 0.03$ & $0.0638 \pm 0.11$ & $0.0388 \pm 0.04$ & $0.0432 \pm 0.05$ & $0.0272 \pm 0.03$ & $0.0228 \pm 0.03$ & $0.0228 \pm 0.03$ \\
Classwise-ECE $\downarrow$ & $0.0315 \pm 0.04$ & $0.0412 \pm 0.04$ & $0.0292 \pm 0.03$ & $0.0394 \pm 0.05$ & $0.0506 \pm 0.04$ & $0.0403 \pm 0.05$ & $0.0221 \pm 0.02$ & $0.0191 \pm 0.02$ & $0.0185 \pm 0.02$ \\
Accuracy $\uparrow$ & $0.9288 \pm 0.10$ & $0.9288 \pm 0.10$ & $0.9240 \pm 0.11$ & $0.9148 \pm 0.12$ & $0.9148 \pm 0.12$ & $0.9083 \pm 0.12$ & $0.9189 \pm 0.11$ & $0.9189 \pm 0.11$ & $0.9111 \pm 0.13$ \\
Macro-F1 $\uparrow$ & $0.8838 \pm 0.09$ & $0.8838 \pm 0.09$ & $0.8661 \pm 0.11$ & $0.8650 \pm 0.11$ & $0.8650 \pm 0.11$ & $0.8397 \pm 0.12$ & $0.8660 \pm 0.11$ & $0.8660 \pm 0.11$ & $0.8386 \pm 0.15$ \\
\bottomrule
\end{tabular}%
}
\end{table*}

%% file: tables/s3_rule_diagnostics.tex
\begin{table*}[htbp]
\centering
\caption{Descriptive rule-level reliability diagnostics by base ensemble. The difference is the probability the rule states minus its precision on the held-out observations assigned to it. No inferential test is attached to these summaries.}
\label{tab:supp-rule-diagnostic}
\small
\resizebox{\textwidth}{!}{%
\begin{tabular}{llrr}
\toprule
Base ensemble & Measure & RC+ & RCProb \\
\midrule
RF & Non-default rules & 12,013 & 6,556 \\
 & Confidence $=1.0$ (\%) & 72.6 & 0.0 \\
 & Confidence $\geq0.999$ (\%) & 74.1 & 8.8 \\
 & Zero held-out assignment (\%) & 38.6 & 31.7 \\
 & \multicolumn{3}{l}{\textit{Mean calibration-in-the-large difference by held-out assignment}} \\
 & 1-9 & 0.1929 & 0.1305 \\
 & 10-29 & 0.0643 & 0.0682 \\
 & 30-99 & 0.0264 & 0.0431 \\
 & $\geq100$ & 0.0048 & 0.0023 \\
\midrule
GBM & Non-default rules & 8,522 & 4,669 \\
 & Confidence $=1.0$ (\%) & 71.5 & 0.0 \\
 & Confidence $\geq0.999$ (\%) & 73.3 & 3.3 \\
 & Zero held-out assignment (\%) & 33.8 & 23.4 \\
 & \multicolumn{3}{l}{\textit{Mean calibration-in-the-large difference by held-out assignment}} \\
 & 1-9 & 0.2076 & 0.1298 \\
 & 10-29 & 0.0785 & 0.0636 \\
 & 30-99 & 0.0306 & 0.0417 \\
 & $\geq100$ & 0.0076 & 0.0039 \\
\bottomrule
\end{tabular}%
}
\end{table*}

%% file: tables/s5_aggregate.tex
\begin{table*}[htbp]
\centering
\caption{Aggregate-inference sensitivity analysis by base ensemble. Within each base ensemble, the same dataset-level Wilcoxon and Holm(12) procedure as the primary decision-list analysis is used.}
\label{tab:supp-aggregate}
\scriptsize
\resizebox{\textwidth}{!}{%
\begin{tabular}{lrrrrrr}
\toprule
 & \multicolumn{3}{c}{RF} & \multicolumn{3}{c}{GBM} \\
\cmidrule(lr){2-4}\cmidrule(lr){5-7}
Measure & RC+Agg & RCProbAgg & $p_{\mathrm{Holm}}$ & RC+Agg & RCProbAgg & $p_{\mathrm{Holm}}$ \\
\midrule
Log-loss $\downarrow$ & 1.6053 & 0.4506 & $<0.0001$ & 1.4278 & 0.4750 & $<0.0001$ \\
Brier score $\downarrow$ & 0.2190 & 0.2090 & 0.9639 & 0.2149 & 0.2066 & 0.6010 \\
Confidence-ECE $\downarrow$ & 0.0850 & 0.0624 & 0.0447 & 0.0773 & 0.0563 & 0.1041 \\
Classwise-ECE $\downarrow$ & 0.0693 & 0.0548 & 0.1166 & 0.0659 & 0.0525 & 0.1982 \\
Macro-F1 $\uparrow$ & 0.8357 & 0.8238 & 0.1795 & 0.8374 & 0.8267 & 0.1982 \\
Fidelity $\uparrow$ & 0.9080 & 0.9131 & 0.9639 & 0.9212 & 0.9107 & 0.0717 \\
Rules $\downarrow$ & 35.92 & 19.50 & 0.0006 & 26.14 & 14.21 & 0.0006 \\
Total antecedents $\downarrow$ & 148.95 & 109.88 & 0.0169 & 97.73 & 71.21 & 0.0007 \\
Distinct antecedents $\downarrow$ & 74.89 & 44.22 & 0.0006 & 56.14 & 32.90 & $<0.0001$ \\
Conditions/rule $\downarrow$ & 3.37 & 4.36 & 0.0117 & 3.07 & 3.82 & 0.1072 \\
\textsc{uniq} $\uparrow$ & 0.636 & 0.568 & 0.0076 & 0.690 & 0.654 & 0.1982 \\
Fit time (s) $\downarrow$ & 12.82 & 13.37 & 0.2575 & 6.74 & 7.45 & 0.1633 \\
\bottomrule
\end{tabular}%
}
\end{table*}

%% file: tables/s6_rule_use.tex
\begin{table*}[htbp]
\centering
\caption{Descriptive held-out use of the extracted rules by base ensemble. A matching rule has an antecedent satisfied by the instance; a selected rule is the first matching rule under decision-list inference. The table describes overlap in the standardized rule representations and is not a comparison of the native inference semantics of the comparison algorithms.}
\label{tab:supp-rule-use}
\scriptsize
\resizebox{\textwidth}{!}{%
\begin{tabular}{llrrrr}
\toprule
Base & Method & Non-default rules & Matches/covered instance & Selected at least once (\%) & Match, never selected (\%) \\
\midrule
RF & RC+ & 34.82 & 4.28 & 68.4 & 27.6 \\
 & RCProb & 19.00 & 5.18 & 73.9 & 25.2 \\
 & DT & 27.83 & 1.00 & 77.2 & 0.0 \\
 & inTrees & 14.89 & 3.68 & 60.6 & 36.7 \\
 & DFT & 3.05 & 1.46 & 88.8 & 10.9 \\
\midrule
GBM & RC+ & 24.70 & 3.16 & 72.1 & 22.7 \\
 & RCProb & 13.53 & 3.38 & 80.1 & 18.8 \\
 & DT & 27.83 & 1.00 & 77.2 & 0.0 \\
 & inTrees & 13.73 & 3.60 & 59.6 & 36.1 \\
 & DFT & 3.10 & 1.50 & 90.2 & 9.2 \\
\bottomrule
\end{tabular}%
}
\end{table*}

%% file: tables/s7_identical_inference.tex
\begin{table}[htbp]
\centering
\caption{Decision-list and aggregate inference in cells where both inference modes selected exactly the same stored ruleset, reported by base ensemble.}
\label{tab:supp-identical-inference}
\small
\resizebox{\textwidth}{!}{%
\begin{tabular}{llrrrrr}
\toprule
Base & Method & Cells & Log-loss (DL) & Log-loss (Agg.) & F1 (DL) & F1 (Agg.) \\
\midrule
RF & RC+ & 233 & 1.700 & 1.424 & 0.840 & 0.841 \\
 & RCProb & 257 & 0.456 & 0.447 & 0.823 & 0.822 \\
\midrule
GBM & RC+ & 221 & 1.475 & 1.231 & 0.853 & 0.854 \\
 & RCProb & 261 & 0.457 & 0.446 & 0.838 & 0.838 \\
\bottomrule
\end{tabular}%
}
\end{table}

%% file: tables/s8_sensitivity.tex
\begin{table*}[htbp]
\centering
\caption{Targeted parameter sensitivity reported separately for RF and GBM. These reduced studies assess stability, not main-benchmark performance. Both panels use a fixed base ensemble over three folds under aggregate inference, and meet at $\tau=5$, $n_0=50$. Panel A varies $n_0$ at $\eta=1$ and $\tau=5$; larger $n_0$ increases log-loss without changing F1 or rule count. Panel B varies $\tau$ at $\eta=1$ and $n_0=50$; results are nearly flat over $\tau\in[0,5]$, and larger values raise RF log-loss and lower GBM macro-F1.}
\label{tab:supp-sensitivity}
\scriptsize
\begin{tabular}{llrrrrrrrr}
\toprule
 & & \multicolumn{4}{c}{RF} & \multicolumn{4}{c}{GBM} \\
\cmidrule(lr){3-6}\cmidrule(lr){7-10}
Setting & Value & Log-loss & ECE & F1 & Rules & Log-loss & ECE & F1 & Rules \\
\midrule
\multicolumn{10}{l}{\textit{A. Mixing support $n_0$}} \\
$n_0$ & 50 & 1.010 & 0.075 & 0.681 & 15.61 & 0.952 & 0.080 & 0.690 & 13.94 \\
$n_0$ & 100 & 1.019 & 0.075 & 0.681 & 15.61 & 0.958 & 0.082 & 0.690 & 13.94 \\
$n_0$ & 200 & 1.029 & 0.076 & 0.681 & 15.61 & 0.964 & 0.081 & 0.690 & 13.94 \\
$n_0$ & 400 & 1.041 & 0.076 & 0.681 & 15.61 & 0.972 & 0.080 & 0.690 & 13.94 \\
\addlinespace
\multicolumn{10}{l}{\textit{B. Shrinkage strength $\tau$}} \\
$\tau$ & 0 & 1.008 & 0.071 & 0.681 & 15.61 & 0.954 & 0.079 & 0.697 & 13.94 \\
$\tau$ & 1 & 1.008 & 0.071 & 0.681 & 15.61 & 0.953 & 0.079 & 0.697 & 13.94 \\
$\tau$ & 2.5 & 1.009 & 0.072 & 0.681 & 15.61 & 0.953 & 0.081 & 0.697 & 13.94 \\
$\tau$ & 5 & 1.010 & 0.075 & 0.681 & 15.61 & 0.952 & 0.080 & 0.690 & 13.94 \\
$\tau$ & 20 & 1.015 & 0.078 & 0.681 & 15.61 & 0.952 & 0.077 & 0.690 & 13.94 \\
$\tau$ & 50 & 1.018 & 0.079 & 0.681 & 15.61 & 0.952 & 0.080 & 0.684 & 13.94 \\
\bottomrule
\end{tabular}
\end{table*}